\documentclass{article}
\usepackage{iclr2027_conference,times}
\newcommand{\buildmode}{preprint} 
\newcommand{\buildsubmission}{submission}
\newcommand{\buildpreprint}{preprint}
\ifx\buildmode\buildsubmission\else\iclrfinalcopy\fi

\usepackage{latexsym}

\usepackage[T1]{fontenc}

\usepackage[utf8]{inputenc}

\usepackage{microtype}

\usepackage{inconsolata}

\usepackage{graphicx}
\usepackage{afterpage}
\usepackage{wrapfig}
\usepackage[font=small,labelfont=bf]{caption}
\usepackage{hyperref}
\hypersetup{colorlinks=true,linkcolor=epidk,citecolor=epidk,urlcolor=epidk} 
\usepackage{etoc} 
\usepackage{placeins} 
\usepackage{url}

\usepackage{booktabs}
\usepackage{makecell} 
\usepackage{amsmath,amssymb}
\usepackage{enumitem}
\usepackage{xcolor}
\definecolor{epi}{HTML}{095792}\definecolor{epidk}{HTML}{0A3F6E}\definecolor{epimid}{HTML}{81A3CF}\definecolor{epilt}{HTML}{CCDFF7}
\definecolor{fail}{HTML}{C85138}\definecolor{faillt}{HTML}{FDDFD7}
\definecolor{mdl}{HTML}{6B7076}\definecolor{mdllt}{HTML}{EEEFF1}\definecolor{ink}{HTML}{222222}
\definecolor{bandA}{HTML}{F2EFCC}\definecolor{bandB}{HTML}{D3C7C0}
\definecolor{epirow}{HTML}{E6EFFA} 
\usepackage{colortbl}
\usepackage{pifont}
\usepackage{todonotes}
\usepackage{tikz}
\usetikzlibrary{arrows.meta,positioning,fit,backgrounds,calc,patterns}
\newcommand{\cmark}{\ding{51}} 
\newcommand{\xmark}{\ding{55}}



\usepackage{xspace}

\newcommand{\methodname}{\textsc{EpiKV}\xspace}
\newcommand{\methodq}{\textsc{EpiKV}\xspace}
\newcommand{\methodseg}{\textsc{EpiKV}$^{-\mathrm{Q}}$\xspace}
\newcommand{\methodflat}{\textsc{EpiKV}$^{-\mathrm{QS}}$\xspace}

\title{\methodname: Epiphany-Aware KV Cache Eviction \\ Without the Attention Matrix}

\author{Steven Kolawole \& Virginia Smith \\
  Carnegie Mellon University \\
  \texttt{\{skolawol,smithv\}@cs.cmu.edu}}

\begin{document}
\maketitle
\ifx\buildmode\buildpreprint\lhead{Preprint. Under review.}\fi
\etocdepthtag.toc{maintag} 

\begin{abstract}
\vspace{-4mm}

Reasoning models can generate chains of thought tens of thousands of tokens long, making the key--value (KV) cache that holds them a major bottleneck for inference throughput. Existing eviction policies for long reasoning traces typically rank cached tokens using \textit{attention weights}, requiring access to the attention matrix and making them incompatible with fast inference kernels. In this work we study the limits of such policies under tight cache budgets. Surprisingly, we find that under the strongest of them the generations that finish are wrong about as often as without eviction; most of the accuracy loss comes from generations that enter loops and run until the length limit, and retaining more tokens according to a fixed importance score exacerbates this behavior. What stops the looping is keeping the tokens the model's recent queries point to, and the forward pass the model already runs reveals them without the attention matrix.
Motivated by this observation, we introduce epiphany-aware KV cache eviction (\methodname), which combines hidden-state shifts with the model's recent query--key relevance to rank cached tokens \textit{without materializing the attention matrix}. On multiple benchmarks, \methodname~matches or outperforms the strongest attention-based eviction baselines while running directly in vLLM with unmodified attention kernels.


\end{abstract}



\vspace{-3mm}
\section{Introduction}
\label{sec:intro}
\vspace{-3mm}

\begin{figure}[b!]
\vspace{-3.5mm}
\centering
\begin{minipage}[c]{0.60\textwidth}
\centering
\includegraphics[width=\linewidth]{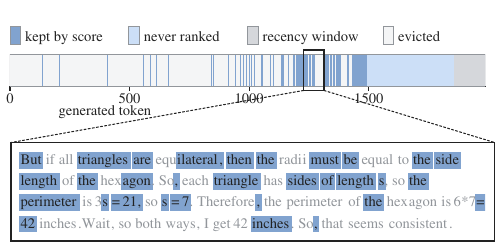}
\end{minipage}\hfill
\begin{minipage}[c]{0.38\textwidth}
\centering
\includegraphics[width=\linewidth]{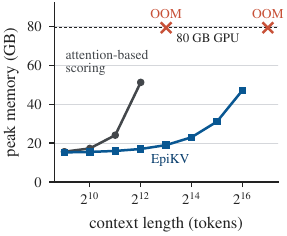}
\end{minipage}
\caption{\textbf{Left: the epiphany score keeps the concluded steps and the insight of a trace
and evicts the connective prose between them.} One MATH-500 trace under \methodseg~at a
1024-token cache: the fate of every generated token, and the words of one stretch; the trace
ends with the correct answer (Appendix~\ref{app:retained}). \textbf{Right: reading the
attention weights exhausts an 80\,GB GPU at 8,192 tokens of prefill; FlashAttention, and so \methodname, fits 65,536.}
Peak memory of one forward pass (\S\ref{sec:r-eng}).}
\label{fig:prefillmem}
\vspace{-2mm}
\end{figure}

Reasoning models such as DeepSeek-R1 \citep{deepseekr1} solve hard problems by
generating long chains of thought; for example, one math problem can take tens
of thousands of tokens of reasoning before the answer. The key--value (KV) cache
grows linearly with this length, which can dominate device memory across concurrent requests~\citep{kvquant} and limits the number of requests a server can
hold~\citep{pagedattention}. KV cache eviction aims to address this by keeping only a budget of $K$ tokens, but raises the question every
eviction method must answer: \emph{which tokens matter?}

Eviction methods built for reasoning traces often answer this via attention weights
\citep{thinkv,raas,longflow}. Attention weight is a noisy proxy for importance~\citep{jain2019attention}:
attention sinks absorb weight whatever their content \citep{streamingllm}, and the
cumulative attention that H2O~\citep{h2o} ranks by tracks counterfactual importance more weakly than
the hidden-state signals we use (\S\ref{sec:signal}). It is
also expensive to read. The weights exist only if the attention matrix is
materialized, which efficient approaches such as FlashAttention are designed to avoid \citep{flashattention,flashattention2}.
Reading them forces the eager kernel, whose memory grows with the number of queries
times the cached tokens: one pass exhausts an 80\,GB GPU at a context shorter than a
single reasoning trace (Figure~\ref{fig:prefillmem}, right), and a server that batches
requests multiplies the queries. No paged serving engine exposes them at all. Without a
custom kernel, the policies built for
reasoning models thus may not be able to run in the engines that serve them. Attention-free
scores exist, from the norm of the cached key \citep{knorm} to block statistics inside a paged
cache \citep{pagedeviction}, but they are fixed when a token is cached and blind to what the
model is still using; the one we test, a fixed-budget form of LagKV \citep{lagkv}, trails at tight budgets (\S\ref{sec:r-acc}).

To see what eviction costs in accuracy, we split every policy's errors on the same
problems into generations that stopped with a wrong answer and generations that never
stopped. Under RaaS~\citep{raas}, the strongest attention-based policy, and under our own policies,
the generations that finish are wrong about as often as without eviction. Instead, we find that the
extra errors are generations that loop until the length limit and never answer, under
every policy and on all models we test (\S\ref{sec:r-failure}), a failure earlier work has observed under
eviction \citep{raas,vase}. On one problem, a policy that ignores the
model's recent queries lets it reach the right answer, restate it hundreds of
times and write the digit 2 until the limit (Figure~\ref{fig:loop}). Keeping
more tokens does not help. Filling the cache with the next-best tokens by a score fixed
when each token was generated makes the looping worse. By contrast, keeping the tokens
the model's recent queries point to lets the generation stop. Which tokens matter
therefore depends on which tokens the model is still using, and the model's own
queries, read against the cached keys, identify them without the attention matrix.

This insight leads us to propose \emph{epiphany-aware} KV cache eviction (\methodname), which ranks tokens from the forward pass the model already runs. \methodname scores each token once,
when the token is generated, by how sharply it shifts the model's hidden states. The
largest shifts mark an \textit{epiphany} in the trace---turning points in the reasoning, such as a concluded step or a
committed insight (Figure~\ref{fig:prefillmem}, left).
Each time \methodname evicts, it also compares the model's own recent queries against the
cached keys, which tells it which cached tokens the model is still using. Neither step needs
the attention matrix, a custom kernel, or training (Figure~\ref{fig:pipeline}). We summarize our contributions below.
\begin{enumerate}[leftmargin=1.4em,itemsep=1pt,topsep=2pt]
  \item \textbf{Where eviction loses accuracy.} On both models and the same problems, we
    identify that the errors eviction adds at tight budgets are generations that never stop:
    most of them under every policy we test and nearly all under the strongest; filling the
    cache by a fixed score raises their number (\S\ref{sec:r-failure}, Table~\ref{tab:termination}, Figure~\ref{fig:loop}).
    The split costs nothing to compute, and we recommend that eviction papers report it
    beside accuracy (\S\ref{sec:discussion}); the marked traces are released with
    the code.
  \item \textbf{An eviction policy without the attention matrix.} On MATH-500, \methodname~matches the
    strongest attention-based policy at every budget on Llama-8B and exceeds every
    baseline at every budget on Qwen-7B; most of the problems its query term wins back
    are generations that now stop (\S\ref{sec:r-failure}, Table~\ref{tab:termination}). On
    AIME 2024--2026 and GPQA-Diamond it is within paired noise of no eviction
    (Appendix~\ref{app:fullresults}). It runs
    inside vLLM~\citep{pagedattention} on stock kernels with the accuracy it has in our harness
    (\S\ref{sec:r-eng}, Table~\ref{tab:vllm-evict}).
  \item \textbf{The signal.} We find that hidden-state change in a mid band of layers marks a
    token's importance and change in an upper-mid band marks its absence, each tracking
    counterfactual importance more closely than cumulative attention, and that a causal rolling
    $z$-score removes the drift the residual stream carries over a long generation
    (\S\ref{sec:signal}). The labels are released with the code.
  \item \textbf{Where reading the attention weights costs.} Crucially, \methodname~never reads the
    attention weights, so it runs on the kernels that serving engines use. We measure what reading
    them costs. It is little when one request decodes at a time, but at prefill and under batching the
    pass that returns them is slower and runs out of memory first (\S\ref{sec:r-eng},
    Figure~\ref{fig:prefillmem}), and a policy that needs them cannot run in a paged serving
    engine without a custom kernel. Serving is where eviction matters most.
\end{enumerate}

\section{Method}
\label{sec:method}
\vspace{-2.5mm}

\subsection{Problem setup}
\label{sec:setup}
\vspace{-1mm}

During autoregressive decoding the key--value (KV) cache grows by one entry per
layer for every generated token. For a trace of $n$ tokens over a model with $L$
layers and $H$ key--value heads of dimension $d_h$, the cache holds $2 L H d_h n$
scalars, and for traces of $10^4$ to $10^5$ tokens it dominates device memory.
Decode-time eviction caps the cache at a budget of $K$ tokens: at each compaction a policy
scores the cached positions, keeps the $K$ highest-scoring ones and discards the rest
permanently. A retained token keeps its \emph{logical} position: its position id and
rotary offset~\citep{su2024roformer} continue from the original index, so eviction changes what the model
attends to and not where the model believes it is. The alternative convention
re-embeds the survivors at the compacted cache's physical slots; we call it the
\emph{rewound} convention. Several public implementations use it~\citep{streamingllm}, and \S\ref{sec:r-failure} and Appendix~\ref{app:convention} show what it does to the results.

\vspace{-1mm}
\begin{figure}[t]
\vspace{-4mm}
\centering
\begin{tikzpicture}[x=1cm,y=1cm,font=\footnotesize,
  >={Stealth[length=1.5mm,width=1.2mm]},
  ttl/.style={font=\footnotesize\bfseries,text=ink,anchor=base west,inner sep=0},
  sub/.style={font=\scriptsize,text=mdl,anchor=base west,inner sep=0,align=left},
  box/.style={draw=mdl,fill=mdllt,rounded corners=1pt,align=center,inner sep=2.2pt,font=\scriptsize,text=ink},
  ours/.style={draw=epi,fill=epilt,rounded corners=1pt,align=center,inner sep=2.2pt,font=\scriptsize,text=ink},
  arr/.style={->,draw=mdl,line width=0.5pt},
  earr/.style={->,draw=epi,line width=0.6pt},
  cell/.style={draw=mdl!60,line width=0.3pt,minimum width=1.3mm,minimum height=3.2mm,inner sep=0,anchor=south west},
  kept/.style={cell,fill=epimid,draw=white,line width=0.4pt},
  filled/.style={cell,pattern=north east lines,pattern color=epi,draw=epi,line width=0.3pt},
]
\draw[mdl!35,line width=0.4pt] (2.92,-0.45) -- (2.92,3.35);
\draw[mdl!35,line width=0.4pt] (7.0,-0.45) -- (7.0,3.35);
\node[ttl] at (0,3.2) {1\;\;One decode step};
\node[sub] at (0,2.88) {unmodified FlashAttention,};
\node[sub] at (0,2.6) {no attention matrix};
\foreach \i in {0,...,6} {\draw[mdl,fill=white,line width=0.4pt] (0.45,0.3+0.3*\i) rectangle (1.65,0.5+0.3*\i);}
\def\yA{1.0}\def\yB{1.9}
\draw[epi,fill=bandA,line width=0.7pt] (0.45,\yA-0.1) rectangle (1.65,\yA+0.1);
\draw[epi,fill=bandB,line width=0.7pt] (0.45,\yB-0.1) rectangle (1.65,\yB+0.1);
\node[font=\scriptsize,text=epi,anchor=east] at (0.42,\yA) {A};
\node[font=\scriptsize,text=epi,anchor=east] at (0.42,\yB) {B};
\node[font=\scriptsize,text=mdl,anchor=north] at (1.05,-0.02) {token $t$};
\draw[arr] (1.05,-0.02) -- (1.05,0.3);
\node[ttl] at (3.05,3.2) {2\;\;Score, once per token};
\node[box,minimum width=15mm] (gB) at (3.95,\yB) {$z\big(\lVert\Delta h_B\rVert\big)$};
\node[box,minimum width=15mm] (gA) at (3.95,\yA) {$z\big(\lVert\Delta h_A\rVert\big)$};
\node[ours,minimum width=20mm] (s) at (5.9,1.45) {$s(t)=z_A-z_B$};
\node[font=\scriptsize,text=mdl,anchor=north] at (5.9,1.18) {fixed at generation};
\node[ours] (h) at (4.9,0.15) {store $h_A$ of the last $w$ tokens\\to recompute their queries};
\draw[earr] (1.65,\yB) -- (gB.west);
\draw[earr] (1.65,\yA) -- (gA.west);
\draw[earr] (2.3,\yA) |- (h.west);
\draw[arr] (gB.east) -- ++(0.15,0) |- ([yshift=2pt]s.west);
\draw[arr] (gA.east) -- ++(0.15,0) |- ([yshift=-2pt]s.west);
\node[ttl] at (7.2,3.2) {3\;\;Compaction};
\node[sub] at (9.5,3.2) {every $\tau$ steps, once the cache exceeds $K$};
\def\ytop{2.05}
\def\ybot{0.3}
\foreach \y in {\ytop,\ybot} {
  \draw[mdl,fill=mdllt,line width=0.3pt] (7.2,\y) rectangle (7.95,\y+0.32);
  \node[font=\scriptsize,text=mdl] at (7.575,\y+0.16) {prompt};
  \draw[mdl,fill=mdllt,line width=0.3pt] (13.1,\y) rectangle (13.85,\y+0.32);
  \node[font=\scriptsize,text=mdl] at (13.475,\y+0.16) {recent};
  \foreach \k in {0,...,3} {\foreach \c in {0,...,7} {\node[cell,fill=white] at (8.08+\k*1.25+\c*0.13,\y) {};}}
}
\foreach \k/\cap in {0/64,1/8,2/32,3/64} {\node[font=\scriptsize,text=mdl] at (8.6+\k*1.25,\ytop-0.17) {cap \cap};}
\foreach \k/\c/\o/\hh in {0/2/13.25/0.92,1/5/13.37/0.76,2/3/13.49/0.6,3/6/13.61/0.46} {
  \draw[epi,line width=0.5pt,->] (\o,\ytop+0.34) .. controls (\o,\ytop+\hh) and (8.145+\k*1.25+\c*0.13,\ytop+\hh) .. (8.145+\k*1.25+\c*0.13,\ytop+0.36);
}
\node[font=\scriptsize,text=epi,anchor=base] at (10.3,2.92) {the last $w$ queries $\cdot$ cached keys $\;\Rightarrow\; r(j)$};
\foreach \k/\c in {0/1,0/2,0/4,0/6, 1/5, 2/3,2/7, 3/0,3/3,3/6,3/7} {\node[kept] at (8.08+\k*1.25+\c*0.13,\ybot) {};}
\foreach \k/\c in {1/2, 1/7, 2/0} {\node[filled] at (8.08+\k*1.25+\c*0.13,\ybot) {};}
\draw[arr,line width=0.6pt] (7.575,\ytop-0.06) -- (7.575,\ybot+0.38);
\node[font=\scriptsize,text=ink,anchor=west,align=left] at (8.0,1.2) {%
  \textbf{tier}: cap each segment by its cached-key variance\\
  \textcolor{epi}{\textbf{rank}: keep the tokens with the highest $r(j)$}\\
  \textcolor{epi}{\textbf{fill}: spend the unused budget by $r(j)$}};
\def\yk{-0.42}
\node[kept] at (8.08,\yk) {};
\node[font=\scriptsize,text=mdl,anchor=west] at (8.24,\yk+0.16) {kept by rank};
\node[filled] at (9.85,\yk) {};
\node[font=\scriptsize,text=mdl,anchor=west] at (10.01,\yk+0.16) {kept by fill};
\node[cell,fill=white] at (11.55,\yk) {};
\node[font=\scriptsize,text=mdl,anchor=west] at (11.71,\yk+0.16) {evicted};
\end{tikzpicture}
\caption{\textbf{\methodname~reads what it needs from the forward pass; the attention matrix
is never formed.} (1)~Decoding runs the unmodified FlashAttention kernel. (2)~Each token gets a
score $s(t)$, fixed at generation, from the hidden-state change at two layers. (3)~Every $\tau$
steps over budget, segments are capped by tier and ranked by $r(j)$, the attention of the last $w$
queries recomputed from Band-A states (\methodseg~ranks by $s(t)$); positions are unchanged.
Blue: what \methodname~adds.}
\vspace{-5mm}
\label{fig:pipeline}
\end{figure}

We call a policy \emph{FlashAttention-compatible} (FA2-compatible) if it computes its
score from two sources only: the cached keys and values, which already reside in
high-bandwidth memory, and the per-layer hidden states that the standard
\texttt{output\_hidden\_states} interface exposes. Such a policy never materializes
the $n \times n$ attention matrix, so it runs inside a FlashAttention forward pass.
A policy is \emph{attention-based} if it needs that matrix. Reading the matrix
disables FlashAttention's tiling and brings back $\mathcal{O}(n^2)$ peak memory for
scoring (Figure~\ref{fig:prefillmem}, right).

\subsection{The epiphany score}
\label{sec:signal}
\vspace{-1mm}

Our signal is the per-token change in the residual stream. For layer $l$ and decode
position $t$, let $h_l(t)$ be the hidden state and define the L2 difference
\begin{equation}
  g_l(t) = \lVert h_l(t) - h_l(t-1) \rVert_2 .
\end{equation}
A large $g_l(t)$ means that generating token $t$ shifted the model's internal state at
layer $l$. Such shifts mark consequential tokens, a committed step or an intermediate
result, whereas fluent filler barely moves it; we call these turning points
\emph{epiphany} tokens and locate them below.

\begin{wrapfigure}{r}{0.41\textwidth}
\vspace{-1.3em}
\centering
\includegraphics{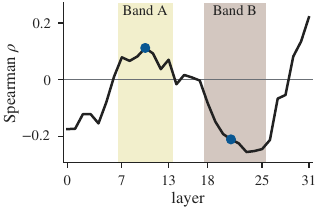}
\caption{\textbf{Hidden-state change marks importance in Band A and its absence in
Band B.} Spearman $\rho$ with counterfactual importance per layer; Llama-8B, MATH-500.
Dots mark the layers the score reads; all layers and AIME-2024 are in
Table~\ref{tab:perlayer}.}
\label{fig:layers}
\vspace{-1.5em}
\end{wrapfigure}
\paragraph{Measuring importance.}
\label{sec:labels}
To find which layers carry this signal we need a ground truth for a token's
importance, and we obtain it by counterfactual occlusion
(Appendix~\ref{app:labeldetails}). For each correctly answered trace we slide a
32-token window with stride 16 over the reasoning span, overwrite the window,
regenerate the answer from the modified context, and label the window important if
the answer changes. Every occlusion feeds the model the same context length, so the
label measures content rather than position. The label is \emph{recoverability under
occlusion}: whether the model still reaches the answer with that span removed, which
is the quantity eviction depends on at decode time. We label with the end-of-sequence
filler and with three stop-free fillers, which agree with one another.

\vspace{-1.5mm}
\paragraph{The two-band anatomy.}
Against these labels, two bands of layers behave consistently and in opposite
directions on competition mathematics (Figure~\ref{fig:layers}). In \emph{Band A}
(layers 7--13) the Spearman $\rho$ between $g_l$ and importance is positive at every
layer, so a high $g_l$ marks an important token. In \emph{Band B} (layers 18--25) it is
negative at every layer, so a high $g_l$ marks a dispensable token. The anatomy holds
under both attention back-ends and under every measured filler, with smaller magnitudes under
the stop-free ones (Appendix~\ref{app:labeldetails}); on AIME-2024 each band keeps
its sign in the mean, with too few labeled traces for a per-layer claim
(Appendix~\ref{app:power}). Each deployed band layer correlates with importance more
strongly than cumulative attention, the signal H2O ranks by, on both configurations
where attention weights exist (Table~\ref{tab:attnsignal}). We interpret the two bands, and how they move with the task, in
\S\ref{sec:discussion}.\footnote{The bands and the window are located once, on MATH-500, and applied
unchanged to every other benchmark. Moving the band layers or the window changes
accuracy within paired noise (Table~\ref{tab:bands}).}

\vspace{-1.5mm}
\paragraph{Positional drift in the residual stream.}
The magnitude of $g_l$ is not stationary over a generation. Within a single trace,
$g_{10}$ drifts downward with position and $g_{21}$ drifts upward, whatever the
tokens contribute. A score read from absolute magnitude therefore tracks position as
much as content, ranking early, droppable tokens above late, load-bearing ones.
Every hidden-state signal we measure carries the drift,
so it belongs to the residual stream and not to a scoring rule. A correlation with
importance labels pooled over whole traces overstates how well such a
signal ranks tokens \emph{within} a trace.

Our score removes the drift by reading \emph{local deviation} instead of magnitude.
Each layer's difference is standardized against its own recent history by a causal
rolling $z$-score,
\begin{equation}
  z_l(t) = \frac{g_l(t) - \mu_l(t)}{\sigma_l(t) + \varepsilon},
  \label{eq:zscore}
\end{equation}
where $\mu_l(t)$ and $\sigma_l(t)$ are the mean and standard deviation of $g_l$ over
the trailing window of $W{=}64$ tokens ending at $t$, $t$ included, and $\varepsilon{=}10^{-6}$;
before the window fills, the statistics use the tokens generated so far, and the first
decode token scores zero. The two bands combine into
\begin{equation}
  s(t) = z_{10}(t) - z_{21}(t).
  \label{eq:combined}
\end{equation}
The window is causal, so a token's score depends only on positions up to $t$; tokens
with high $s$ are retained. The standardization adapts the
lag-relative normalization of \citet{lagkv} and the positional-bias correction of
\citet{ahakv} to hidden-state differences, and averaging over its window raises the correlation of
every signal we deploy above its raw form (Table~\ref{tab:smoothing}).

\subsection{\methodq}
\label{sec:policies}
\vspace{-1.5mm}

\methodq~holds a reasoning trace in a cache of $K$ tokens by making three decisions
at each compaction, all from tensors the forward pass already produces: how much of
each 128-slot \emph{segment} of the cache to keep, which tokens within a segment the
model is still \emph{reading}, and how to \emph{spend} the budget the first decision
leaves over. The second decision governs whether a generation stops
(\S\ref{sec:r-failure}). The score $s(t)$ of Eq.~\eqref{eq:combined} is computed when
token $t$ is generated and never revisited. Every $\tau{=}128$ decode steps, if the
cache exceeds $K$, one compaction applies the three decisions and rewrites the cache.
The prompt tokens and a window of the 128 most recent tokens are always
kept.\footnote{Methods differ in which structural tokens they preserve (attention
sinks, a recency window, the prefill). Appendix~\ref{app:policies} tabulates each
method's choice.}

\vspace{-1.5mm}
\paragraph{Segments and tiers.} At each compaction the cache's decode span is cut into
consecutive 128-slot segments. Each segment is classified as reasoning, execution or
transition by the tertile of its mean key variance over the head dimension, the
attention-free counterpart of ThinKV's segment sparsity, and is assigned a cap of 64, 32
or 8 retained tokens accordingly. Appendix~\ref{app:segablation} compares two alternative segment signals,
the entropy of that variance and ThinKV's attention sparsity itself.

\vspace{-1.5mm}
\paragraph{The query term.} Within a segment, tokens are ranked by how much the
model's recent queries attend to them, read once per compaction and without the
attention matrix. Let $t_1<\dots<t_w$ be the last $w{=}8$ generated tokens and $\ell$
the layer after the score's Band-A layer (layer 11 on Llama-8B, layer 21 on Qwen-7B). From the Band-A
hidden state stored for each $t_i$, which is that layer's input, we recompute its query,
$q_h(t_i) = \mathrm{RoPE}_{t_i}\!\big(W^{q}_{h}\, \mathrm{LN}(h_{\ell-1}(t_i))\big)$,
and dot it with the cached keys of layer $\ell$:
\begin{equation}
  r(j) = \frac{1}{wH}\sum_{i=1}^{w}\sum_{h=1}^{H}
  \mathrm{softmax}_{j' \in \mathcal{C},\, j' \le t_i}\!\Big(\tfrac{q_h(t_i)\cdot k_h(j')}{\sqrt{d_h}}\Big)_{j},
  \label{eq:relevance}
\end{equation}
a causal softmax over the cached positions $\mathcal{C}$ for each query and head, averaged over
both. The relevance $r$ is $z$-scored over the cache and ranks the tokens within each
segment. Computing it is one matrix product of $w$ queries against the cached keys per
head, on keys the compaction already gathers, once every 128 steps. Nothing is
materialized inside the forward pass, and the kernel is untouched. The policy needs only
the last $w$ Band-A hidden states, two rolling buffers of $W$ values and one
score per cached token, a few hundred kilobytes for an 8B model.

\begin{wraptable}{r}{0.55\textwidth}
\vspace{-1.2em}
\centering
\caption{\textbf{Only our policies run on stock FlashAttention; every attention-based
baseline needs the eager kernel or its own.} \emph{Eager} materializes the attention
weights; \emph{custom} needs a bespoke kernel, a serving-engine fork included;
\emph{stock} runs on unmodified FlashAttention-2. $^{\dagger}$A causal variant of
\citet{lagkv}.}
\label{tab:methods}
\scriptsize
\setlength{\tabcolsep}{2.5pt}
\begin{tabular}{@{}lll@{}}
\toprule
Method & Signal & Kernel \\
\midrule
\multicolumn{3}{@{}l}{\emph{Attention-based baselines}}\\
H2O      & cumulative attention & \xmark~eager \\
ThinKV   & attention sparsity per segment & \xmark~custom \\
RaaS     & time each token was last attended & \xmark~eager \\
R-KV     & attention $-$ key redundancy & \xmark~eager \\
LongFlow & attention $\times \lVert v\rVert_1$ & \xmark~custom \\
\midrule
\multicolumn{3}{@{}l}{\emph{Ours: \methodname~and its ablations}}\\
\rowcolor{epirow}\methodq      & tiers, query term $r(j)$, fill & \cmark~stock \\
\methodseg    & tiers, ranked by $s(t)$ & \cmark~stock \\
\methodflat   & global top-$K$ of $s(t)$ & \cmark~stock \\
raw score     & global top-$K$ of $\bar{g}_{10}-\bar{g}_{21}$ & \cmark~stock \\
\midrule
\multicolumn{3}{@{}l}{\emph{Ours: KV-vector signals}}\\
KV-key    & key variance & \cmark~stock \\
Lag-KV$^{\dagger}$ & lag-normalized key and value & \cmark~stock \\
\bottomrule
\end{tabular}

\end{wraptable}
\paragraph{The filled budget.} The tier caps can bind before the budget does and leave
part of $K$ unused, up to about half of it on long traces. \methodq~spends that remainder on the evicted tokens
with the highest $r$, so a compaction lands at $K$. Spending it by $s(t)$ instead costs
accuracy and makes more generations loop (Appendix~\ref{app:budgetfill}), because a score
frozen at generation time keeps tokens the model has stopped reading.
The query term and the fill therefore ship together.

\paragraph{Compaction schedule.} The cache is compacted at segment boundaries, every
$\tau{=}128$ decode steps, and only when it is over budget; $\tau{=}1$ is the per-step
form every baseline but R-KV uses. Between boundaries the cache can exceed $K$ by up to
$\tau{-}1$ entries, so the occupancy follows a saw-tooth around the budget
(Figure~\ref{fig:retainedfull} shows it on one trace). The amortized form is the one
a paged engine can run (\S\ref{sec:r-eng}). Appendix~\ref{app:tau} shows that
compacting more often never improves accuracy and leaves the per-token cost unchanged.

\subsection{Ablations and baselines}
\label{sec:baselines}
\vspace{-1mm}

Table~\ref{tab:methods} lists every policy we evaluate. Three ablations of \methodq
remove its parts in turn; \methodseg~appears in every main table and \methodflat~in the
appendix. \methodseg~drops the query term
and the fill: it ranks within segments by $s(t)$ itself and leaves the caps unspent.
\methodflat~drops the segments as well: it is a single global top-$K$ over $s(t)$.
The raw score is \methodflat~without the detrending: a global top-$K$ over
$\bar g_{10}-\bar g_{21}$.

The KV-vector family scores tokens from quantities already in the cache and needs no
hidden states. \emph{KV-key} uses the rolling-mean variance of the key vectors across
the head dimension; the value-vector and key-only lag variants are released with
the code. \emph{Lag-KV} brings the lag-relative normalization of
\citet{lagkv} into a fixed-budget decode policy. Each token's key and value vectors are
normalized by the per-channel range of the \emph{previous} chunk before the variance is
taken, which removes domain-level shifts in magnitude, and every compaction ranks the
whole cache against the budget $K$. LagKV instead normalizes against the \emph{next}
partition, keeps a fixed share of each partition per head and never revisits a
compressed partition, so its cache grows with the sequence.

Every baseline runs in our harness at matched budgets; Appendix~\ref{app:policies} lists
each policy's structural tokens and Appendix~\ref{app:adapted} its adaptations.

\section{Results}
\label{sec:results}
\vspace{-2mm}

\subsection*{Experimental setup}
\vspace{-1.5mm}

\paragraph{Models.}
DeepSeek-R1-Distill-Llama-8B (32 layers) \citep{deepseekr1} is the primary model,
chosen for direct comparability with ThinKV and as an open-weight member of the
reasoning-model class. DeepSeek-R1-Distill-Qwen-7B (28 layers; score layers 20
and 24, Appendix~\ref{app:qwen}) is the second model in every accuracy table.
Generation is greedy throughout, so repeated runs of one configuration agree exactly.

\paragraph{Datasets.}
MATH-500 \citep{hendrycks2021math,math500} is the primary benchmark: competition
mathematics with verifiable boxed answers and traces of several thousand tokens; we
use the 495 problems that have a boxed reference answer. AIME 2024--2026 \citep{matharena}, pooled over the three
years, tests higher cache pressure with traces several times longer. GPQA-Diamond
\citep{gpqa}, graduate-level science as four-way multiple choice, tests a domain outside
mathematics. GSM8K
\citep{gsm8k} serves only as a difficulty-regime probe for the layer anatomy
(Appendix~\ref{app:gsm8k}), not as an accuracy benchmark.

\paragraph{Budgets and metrics.}
The cache budgets are $K \in \{512, 1024, 2048, 4096\}$ on MATH-500,
$\{4096, 8192\}$ on AIME and $\{1024, 2048, 4096\}$ on GPQA-Diamond.
Generation stops at the end-of-sequence token or at a cap of 8,192 tokens on MATH-500
and 16,384 on AIME and GPQA-Diamond. We report accuracy as exact match on the boxed answer, the number
of generations that reach the cap, and wall-clock time and peak GPU memory per problem.

\vspace{-1.5mm}
\paragraph{Attention back-end.}
Our policies run under FlashAttention-2 because they can; the attention-based policies
run under the eager kernel because only it returns the weights they read. The two
kernels compute the same attention, and every row scores the same problems. Every cell of the
main accuracy tables ran on one platform under one library stack, the one on which our
policies score lower (Appendix~\ref{app:impl}); the hardware, the
versions, and the ablations that ran elsewhere are listed there. The remaining results tables are in
Appendix~\ref{app:fullresults}.

\subsection{One read of the query is worth the attention matrix}
\label{sec:r-acc}
\vspace{-2mm}

\methodq~matches the best attention-based policy at every budget on Llama-8B and
exceeds every baseline at every budget on Qwen-7B (Table~\ref{tab:acc-math}). Every
row of the table runs the same problems, so every comparison is paired
(Table~\ref{tab:pairs}). On Llama-8B \methodq~is within paired noise of RaaS and of the uncapped model at every
budget. On Qwen-7B its lead over RaaS is significant from
$K{=}1024$ upward, its lead over every other attention-based row is significant at
every budget, and at the two largest budgets it is within paired noise of the
uncapped model.

\begin{table}[t]
\centering
\caption{\textbf{\methodq~matches the best attention-based policy on Llama-8B and exceeds
every baseline on Qwen-7B, at every budget.} MATH-500 accuracy (\%), $n{=}495$ per cell,
greedy; every row runs the same problems on one platform. Bold: best eviction row in a
column. $^{\dagger}$Above every attention-based row (exact McNemar $p<0.05$; counts in
Table~\ref{tab:pairs}). FA2: runs on FlashAttention. $^{\star}$No eviction is
budget-independent; most of its generations fit in 4,096 tokens (69\% and 74\%).}
\label{tab:acc-math}
\small
\setlength{\tabcolsep}{3pt}
\begin{tabular}{@{}llrrrrrrrr@{}}
\toprule
 & & \multicolumn{4}{c}{Llama-8B} & \multicolumn{4}{c}{Qwen-7B} \\
\cmidrule(lr){3-6}\cmidrule(lr){7-10}
Method & FA2 & $K{=}512$ & 1024 & 2048 & 4096 & $K{=}512$ & 1024 & 2048 & 4096 \\
\midrule
none$^{\star}$ & \cmark & \multicolumn{4}{c}{69.9} & \multicolumn{4}{c}{79.0} \\
\midrule
\rowcolor{epirow}
\methodq~(ours) & \cmark & \textbf{66.5} & \textbf{70.3} & 71.5 & 70.5 & \textbf{66.1} & \textbf{75.2}$^{\dagger}$ & 78.0$^{\dagger}$ & 78.2$^{\dagger}$ \\
\methodseg~(no query term) & \cmark & 57.8 & 63.2 & 68.9 & 71.3 & 63.0 & 70.7 & \textbf{78.2} & \textbf{79.8} \\
Lag-KV & \cmark & 52.7 & 60.4 & 66.9 & 69.7 & 34.5 & 53.5 & 70.9 & 77.6 \\
\midrule
RaaS & \xmark & 64.8 & 69.9 & \textbf{71.9} & 71.7 & 65.3 & 69.9 & 73.1 & 72.5 \\
H2O & \xmark & 56.6 & 66.7 & \textbf{71.9} & 72.3 & 52.5 & 65.3 & 70.5 & 71.3 \\
R-KV & \xmark & 45.1 & 60.2 & 70.5 & \textbf{72.5} & 30.1 & 60.0 & 72.1 & 73.9 \\
LongFlow & \xmark & 44.0 & 62.4 & 70.1 & 72.1 & 33.1 & 58.4 & 71.7 & 72.9 \\
ThinKV & \xmark & 26.9 & 39.2 & 52.7 & 65.5 & 16.6 & 30.5 & 50.5 & 68.5 \\
 
\bottomrule
\end{tabular}
\vspace{-3mm}
\end{table}

The query term recovers the tight budgets. Without it, \methodseg~already leads Lag-KV at
every budget on both models, and on
Qwen-7B it leads the whole field from $K{=}2048$ upward and is level with RaaS at
$K{=}1024$. On Llama-8B, however, RaaS leads it at the two tight
budgets. \methodq's wins over
\methodseg~concentrate there on Llama-8B and at $K{=}1024$ on
Qwen-7B; elsewhere the two policies are within paired noise of each other. On
Llama-8B the whole field converges at the largest budget, because most MATH-500
generations finish under it and eviction rarely fires. On Qwen-7B the attention-based
rows stay below \methodq~even there.

Which attention-based policy sets the bar depends on the budget. At the two tight
budgets RaaS is the strongest attention-based row on both models; at the loose
budgets the attention-based rows converge, and on Llama-8B H2O ties or passes RaaS.
RaaS keeps a token while the
current queries attend to it, so it knows how
relevant each cached token is to what the model is doing now, which a score frozen at
generation time cannot know. RaaS pays for that signal with the attention matrix at
every step. The hidden-state score gives up the query to avoid that cost, and
\methodq~restores it with one read per compaction (\S\ref{sec:policies}), which recovers
RaaS's margin at the tight budgets.

ThinKV is last at every budget on both models, and neither its refresh window nor its
classifier explains its place on its own (Appendices~\ref{app:thinkvfaithful}
and~\ref{app:segablation}). R-KV and LongFlow run on the same
problems and platform as every other row. On Llama-8B both trail RaaS and H2O at every
budget below the largest, on Qwen-7B both trail \methodseg~at every budget, and both
need the attention probabilities at every compaction (Table~\ref{tab:methods}). To
see where the accuracy lost at tight budgets goes, \S\ref{sec:r-failure} splits the
per-problem records by how each generation ended.

\subsection{Accuracy is lost to loops, not wrong answers}
\label{sec:r-failure}
\vspace{-2mm}

\begin{wraptable}{r}{0.57\textwidth}
\vspace{-1.1em}
\centering
\caption{\textbf{Eviction loses accuracy by failing to stop, not by answering wrongly.}
MATH-500 errors at $K{=}1024$, $n{=}495$; a cap hit ran to the 8,192-token cap. Bottom: problems
right under the first policy and wrong under the second; in parentheses, those where the wrong
side hit the cap.}
\label{tab:termination}
\footnotesize
\setlength{\tabcolsep}{2.6pt}
\begin{tabular}{@{}lrrrrrr@{}}
\toprule
 & \multicolumn{3}{c}{Llama-8B} & \multicolumn{3}{c}{Qwen-7B} \\
\cmidrule(lr){2-4}\cmidrule(lr){5-7}
 & errors & \makecell{cap\\hits} & \makecell{term.\\wrong} & errors & \makecell{cap\\hits} & \makecell{term.\\wrong} \\
\midrule
none & 149 & 89 & 60 & 104 & 55 & 49 \\
\rowcolor{epirow}
\methodq & 147 & 91 & 56 & 123 & 73 & 50 \\
\methodseg & 182 & 122 & 60 & 145 & 93 & 52 \\
RaaS & 149 & 87 & 62 & 149 & 99 & 50 \\
H2O & 165 & 97 & 68 & 172 & 124 & 48 \\
ThinKV & 301 & 158 & 143 & 344 & 191 & 153 \\
 
\midrule
\methodq, not \methodseg & \multicolumn{3}{c}{54 (40 looped)} & \multicolumn{3}{c}{38 (29 looped)} \\
\methodseg, not \methodq & \multicolumn{3}{c}{19 (17 looped)} & \multicolumn{3}{c}{16 (12 looped)} \\
RaaS, not none & \multicolumn{3}{c}{42 (32 looped)} & \multicolumn{3}{c}{18 (15 looped)} \\
none, not RaaS & \multicolumn{3}{c}{42 (31 looped)} & \multicolumn{3}{c}{63 (55 looped)} \\
\bottomrule
\end{tabular}
\vspace{-0.8em}
\end{wraptable}
At tight budgets eviction costs accuracy by breaking termination. We split every
error into two kinds. A \emph{cap hit} is a generation that reached the length cap
without an answer, the signature of a model that has begun to repeat itself. A
\emph{terminated wrong answer} is a generation that stopped on its own and answered
wrongly. VaSE traces such loops to the eviction of large-magnitude values \citep{vase};
protecting those values changes neither our loops nor accuracy (Appendix~\ref{app:valreserve}).
Done for every policy on the same problems, the split measures how much of the lost
accuracy the loops account for; Table~\ref{tab:termination} gives it at $K{=}1024$. RaaS, the hidden-state policies
and the uncapped model answer almost the same number of terminated generations
wrongly, and the whole of \methodseg's excess error over the uncapped model is cap
hits. ThinKV is the exception: it adds terminated wrong answers in large
numbers as well as loops. Figure~\ref{fig:loop} shows one loop, in which the model
reaches the right answer and cannot stop.

\begin{figure}[t]
\centering
\footnotesize
\begin{minipage}[t]{0.485\linewidth}
\textbf{\methodseg: cap hit at 8,192 tokens}\par\vspace{2pt}
\colorbox{black!5}{\parbox{0.96\linewidth}{\ttfamily\scriptsize\raggedright
So, 2220 is the smallest four-digit number. [...] So, is 2220 the answer? Wait, but
2220 is 2220, but 2220 is 2220, \textcolor{fail}{which is 2220, which is 2220, which is 2220,}
[... \textcolor{fail}{``which is 2220''} 279 times ...] So, the number is
\textcolor{fail}{22222222222222} [... to the cap]}}
\end{minipage}\hfill
\begin{minipage}[t]{0.485\linewidth}
\textbf{\methodq: stops at 3,505 tokens}\par\vspace{2pt}
\colorbox{black!5}{\parbox{0.96\linewidth}{\ttfamily\scriptsize\raggedright
[...] 2220 is 2220, which is 2220/30 = 74, which is integer. So, 2220 is the first
four-digit number that works. Therefore, 2220 is the smallest positive integer
multiple of 30 that can be written with only the digits 0 and 2. [...]
\textcolor{epi}{\textbackslash boxed\{2220\}}}}
\end{minipage}
\caption{\textbf{The model reaches the answer and cannot stop.} MATH-500 problem 10 (the
smallest multiple of 30 written with only the digits 0 and 2), Llama-8B, $K{=}1024$.
Without the query term the model restates the correct answer, then writes the digit
2 until the cap; with it, it checks the answer and stops (Appendix~\ref{app:loop}).}
\label{fig:loop}
\vspace{-3mm}
\end{figure}

The same accounting explains how an eviction policy can match the uncapped model at
a tight budget. RaaS ties the uncapped model on Llama-8B at $K{=}1024$ with nearly
the same count of terminated wrong answers, and most of the problems the two trade
are generations that loop under one cache and stop under the other
(Table~\ref{tab:termination}). Which generations loop therefore depends on the cache,
and a tie in accuracy can hide dozens of problems that change sides.

The query term of \methodq~acts through this channel and through no other. At
$K{=}1024$ it lowers \methodseg's cap hits on both models and leaves its terminated
wrong answers where they were, and most of the problems it wins from \methodseg~are
generations that hit the cap under \methodseg~and now stop
(Table~\ref{tab:termination}). Its terminated wrong answers match RaaS's and the
uncapped model's, and on the problems the uncapped model solves it loops on the same share
on both models, whereas RaaS loops more often on Qwen-7B (Appendix~\ref{app:loopshare}).

Doubling the generation cap leaves both policies' accuracy within paired noise, and most
generations that reach the shorter cap reach the longer one too (Appendix~\ref{app:capraise}),
so most cap hits are loops that more length would not finish.
Nor is the remedy simply to keep more. Filling \methodseg's cache to budget with the
next-best tokens by the frozen score, everything else fixed, raises its cap-hit count
and lowers its accuracy in most of the cells we ran, on both models; \methodq~spends the same budget
by the query term instead (Appendix~\ref{app:budgetfill}). Accuracy follows
the tokens the model is still reading when it decides to stop.

Under the rewound convention of \S\ref{sec:setup} the failure is total: H2O loops on every
problem at the tightest budget (Appendix~\ref{app:convention}); under the logical convention,
which every main-text number uses, it is competitive at every budget. Non-termination also sets
a policy's speed (\S\ref{sec:r-eng}).

\subsection{Reading the attention weights costs at prefill and under batching}
\label{sec:r-eng}
\vspace{-1mm}

\begin{wraptable}{r}{0.47\textwidth}
\vspace{-1.1em}
\centering
\caption{\textbf{Both policies keep their harness accuracy when they evict inside vLLM on its
stock kernels.} MATH-500 problems 0--99, Llama-8B, greedy; no paired vLLM--harness difference is
significant.}
\label{tab:vllm-evict}
\footnotesize
\setlength{\tabcolsep}{3pt}
\begin{tabular}{@{}lrrrr@{}}
\toprule
 & \multicolumn{2}{c}{$K{=}1024$} & \multicolumn{2}{c}{$K{=}2048$} \\
\cmidrule(lr){2-3}\cmidrule(lr){4-5}
 & \makecell{acc.\\(\%)} & \makecell{cap\\hits} & \makecell{acc.\\(\%)} & \makecell{cap\\hits} \\
\midrule
\methodseg, vLLM & 68.0 & 24 & 74.0 & 17 \\
\methodseg, harness & 66.0 & 23 & 73.0 & 18 \\
\midrule
\rowcolor{epirow}\methodq, vLLM & 75.0 & 17 & 75.0 & 15 \\
\rowcolor{epirow}\methodq, harness & 76.0 & 16 & 76.0 & 18 \\
\bottomrule
\end{tabular}
\end{wraptable}
\paragraph{One request at a time.}
An attention-based policy reads the attention weights, one number for every pair of a
query token and a cached token. When one request decodes at a time, each step has a
single query token, so the weights are small and reading them is cheap; the cost of
the step is set by loading the model's weights and launching its kernels
\citep{pope2023scaling,flashdecoding,memorybound}. Measured directly, a decode step that returns the
weights takes the same time as a FlashAttention step at every cache length up to our
largest budget (Table~\ref{tab:decode-step}, Appendix~\ref{app:prefill}). An end-to-end run of every
policy at equal load agrees. The attention-based policies are among the fastest, and a
policy's speed follows the work it does between forward passes; \methodq's query
term, one product per compaction, costs nothing measurable per token
(Table~\ref{tab:time-aime}). The time a problem takes follows termination instead.
A loop emits tokens until the cap, so the loops of
\S\ref{sec:r-failure} cost wall-clock time too (Table~\ref{tab:tau}).

\vspace{-1.5mm}
\paragraph{Prefill and batching.}
At prefill every prompt token is a query, and under batching every request adds one,
so the weights grow with the number of queries times the cache. There the pass that
returns them is slower than a FlashAttention pass and runs out of memory at lengths
and batch sizes that FlashAttention serves (Figure~\ref{fig:prefillmem},
Appendix~\ref{app:prefill}). A server runs in this regime, and there a budget
matters for a second reason. It holds each request's cache flat as traces grow, and
the size of that cache decides how many requests fit (Appendix~\ref{app:throughput}).

\vspace{-1.5mm}
\paragraph{Serving, where no engine exposes the weights.}
No vLLM decode path exposes the attention weights, so attention-based scoring cannot run there
without forking the paged-attention kernel. Both of our policies run there on stock kernels, with
no change to the engine beyond Python hooks in the scheduler's block allocator, which tell it
which blocks a compacted request still owns (Table~\ref{tab:vllm-evict}). The port compacts the
paged cache in place; the attention kernels, the paged-cache layout and the engine loop are untouched,
and no CUDA is written.

The port reaches the harness's accuracy. \methodq's query term is read from vLLM's own
fused projection and rotary module, and its ranking differs from the harness's in detail
(Appendix~\ref{app:vllm}). The cost is small: scoring inside the engine takes a modest share of
decode throughput at every concurrency we ran, with negligible tracker state
(Table~\ref{tab:vllm}); capping the cache takes a similar share while KV memory is plentiful, and
once the uncapped engine's KV pool is full the two engines decode at about the same rate
(Figure~\ref{fig:vllm-tp}, Appendix~\ref{app:vllm}).

\section{Related Work}
\label{sec:related}
\vspace{-2mm}

\paragraph{Attention-based eviction.}
Most eviction policies rank tokens by attention weight, whether they target long
prompts \citep{h2o,streamingllm,snapkv,pyramidkv,chunkkv,caote,ahakv,dynamickv} or the long generations of
reasoning models \citep{thinkv,raas,rkv,longflow,foresightkv}.
Table~\ref{tab:methods} lists what each policy we compare against reads and the kernel
it needs. Reading attention weights requires the attention matrix that fused kernels
such as FlashAttention avoid forming \citep{flashattention,flashattention2}, so every
attention-based baseline runs on the eager kernel or its own.

\vspace{-1.5mm}
\paragraph{Eviction-induced loops.}
Eviction can send a reasoning model into repetition. RaaS reports it for H2O
\citep{raas}, and VaSE attributes it to evicting large-magnitude value states and
protects them \citep{vase}. These works establish that the loops occur; the split of \S\ref{sec:r-failure} shows,
further, that at tight budgets they account for the accuracy lost.

\vspace{-1.5mm}
\paragraph{Signals without the attention matrix.}
Attention-free scores exist: the $\ell_2$ norm of the cached key \citep{knorm}, block-level
statistics inside vLLM's paged cache \citep{pagedeviction}, and lag-relative KV statistics
\citep{lagkv}. All are fixed when a token is cached and blind to the model's current queries.
LagKV keeps a fixed share of each partition, so its cache grows with the sequence; our
Lag-KV brings its normalization into a fixed-budget policy (\S\ref{sec:baselines}) and
represents this family in our comparison; it trails the query-aware policies at tight budgets
(Table~\ref{tab:acc-math}). Outside eviction, speculative decoding already drafts from hidden states because
they carry more predictive structure than token sequences \citep{eagle}. Within eviction, no prior decode-time
policy for reasoning traces scores tokens from the residual stream or recomputes the
query from it.

\vspace{-1.5mm}
\paragraph{Orthogonal directions.}
Retrieval keeps every token and fetches a subset per step \citep{quest,freekv}, SideQuest
trains the model to delete stale tool output \citep{sidequest}, and quantization lowers
the precision of what is kept \citep{kvquant,minikv}; each composes with eviction.

\section{Discussion}
\label{sec:discussion}
\vspace{-2mm}

\paragraph{Band A composes content; Band B predicts the next token.}
Band A coincides with the mid-network layers that interpretability work identifies
as the site of factual recall and feature composition \citep{rome,memit,lad2025stages}, so
a large hidden-state change there marks a token at which the model retrieves or
composes content. Band B sits where the residual stream turns toward the output
distribution \citep{geva2021,lad2025stages}. A large change there accompanies fluent, predictable continuation, and
the tokens it marks are the ones the model can do without. Subtracting the bands uses
both signals at once; at eviction time the difference and Band A alone are within noise
(Appendix~\ref{app:bands}). The bands also move with the task, deeper on competition
mathematics than on GSM8K (Appendix~\ref{app:gsm8k}), and the residual stream drifts
with position over a long generation (\S\ref{sec:signal}). Any score read from the
stream inherits that drift; a causal rolling $z$-score is a cheap remedy.

\vspace{-1.5mm}
\paragraph{Report loops and wrong answers separately.}
Accuracy alone hides how a policy fails. The same loss can come from loops or from
wrong answers, and the two call for different fixes. \methodseg's loss comes from loops
alone, for instance, and ThinKV's from both (Table~\ref{tab:termination}). The split
costs nothing to compute from the per-problem records every harness already writes,
and it tells a designer which lever moved. We recommend that eviction papers report it
beside accuracy, together with the generation cap, since the cap decides how a loop is
counted.

\section{Conclusion}
\label{sec:conclusion}
\vspace{-2mm}
At tight budgets, the accuracy that eviction costs a reasoning model goes to
generations that never stop. Keeping the tokens the model's recent queries point to
lets them stop, and \methodname~identifies those tokens from the cached keys and the hidden
states the forward pass already returns. Because it needs neither the attention matrix
nor a custom kernel, \methodname~runs where reasoning models are served: inside a
paged engine, on FlashAttention, at the tight budgets where eviction matters most. The
split of errors that located the problem also gives the next policies their target, a
cache that keeps every generation able to stop.

\ifx\buildmode\buildsubmission\else
\section*{Limitations}
\label{sec:limitations}
\addcontentsline{toc}{section}{Limitations and future work}
Both models are distilled from DeepSeek-R1. Every result uses greedy decoding, which
keeps every comparison between policies exactly paired. On AIME and GPQA-Diamond the uncapped
model itself fails to stop on a large share of problems, so those benchmarks bound what any policy
can recover (Tables~\ref{tab:acc-aime} and~\ref{tab:gpqa}). Kernel costs and
serving throughput are measured on a single GPU.

\paragraph{Future work.}
The query term reads one layer from a short window of recent queries; a deeper read, or
a relevance term inside the tiering as well, is the natural next step toward the
uncapped model at the tightest budget. Beyond occlusion, in-distribution replacement
and gradient-based attribution would test whether counterfactual importance is a stable
target; the labels we release under every filler are the baseline for that comparison.

\fi

\subsection*{Reproducibility statement}
The harness, every eviction policy including the adapted baselines
(Appendix~\ref{app:adapted}), the counterfactual labels under every filler
(\S\ref{sec:labels}), the marked traces behind Figure~\ref{fig:prefillmem} (left), the
kernel-cost microbenchmarks (Appendix~\ref{app:prefill}) and the vLLM integration are
released with the code. The experimental setup at the start of \S\ref{sec:results} and
\S\ref{sec:policies} give every setting of \methodq~and its ablations, and
Appendix~\ref{app:impl} lists the hardware and library stack behind each table. Greedy
decoding makes each configuration deterministic on its platform, so a rerun on the same
stack reproduces every number exactly, and every comparison between policies is made
within one stack (Appendix~\ref{app:impl}).

\subsection*{AI use statement}
In this work, we used generative AI tools as coding and writing assistants: to implement
and run experiment and analysis code, and to edit prose, under the authors' direction.
The research questions, the method and the claims are the authors'. The authors reviewed
and tested all AI-assisted code, checked every reported number against the underlying
result files, and take full responsibility for the content of this paper.

\ifx\buildmode\buildsubmission\else
\subsubsection*{Acknowledgments}
We thank Vashisth Tiwari for their helpful comments and pointers in the ideation of this work.
\fi

\bibliography{custom}
\bibliographystyle{iclr2027_conference}

\appendix
\clearpage
\etocdepthtag.toc{appendixtag}
\etocsettagdepth{maintag}{none}
\etocsettagdepth{appendixtag}{subsection}
\begingroup
\hypersetup{pdfborder={0 0 0}}
\etocsettocstyle{\noindent{\large\bfseries Appendix contents}\par\vspace{0.8em}}{\vspace{1.2em}}
\tableofcontents
\ifx\buildmode\buildsubmission
\section*{Limitations}
\label{sec:limitations}
\addcontentsline{toc}{section}{Limitations and future work}
Both models are distilled from DeepSeek-R1. Every result uses greedy decoding, which
keeps every comparison between policies exactly paired. On AIME and GPQA-Diamond the uncapped
model itself fails to stop on a large share of problems, so those benchmarks bound what any policy
can recover (Tables~\ref{tab:acc-aime} and~\ref{tab:gpqa}). Kernel costs and
serving throughput are measured on a single GPU.

\paragraph{Future work.}
The query term reads one layer from a short window of recent queries; a deeper read, or
a relevance term inside the tiering as well, is the natural next step toward the
uncapped model at the tightest budget. Beyond occlusion, in-distribution replacement
and gradient-based attribution would test whether counterfactual importance is a stable
target; the labels we release under every filler are the baseline for that comparison.

\fi
\endgroup

\FloatBarrier
\section{Harness, platforms and baselines}
\label{app:g-setup}

\FloatBarrier
\subsection{Harness and platforms}
\label{app:impl}

Eviction is applied through the HuggingFace \texttt{DynamicCache}. Retained tensors
are written back into the live cache's own per-layer slots rather than into a freshly
constructed cache; a policy that evicts at every step would otherwise hold two copies
of the cache, a cost that grows with the cache. \texttt{\_seen\_tokens}
is set to the retained length, without which the model builds a causal mask one
position too long. Keep-masks are placed on each tensor's own device, so multi-GPU
\texttt{device\_map="auto"} runs are correct. FlashAttention runs use a single GPU. The
no-eviction baseline is run once and reused across budgets. Table~\ref{tab:hosts}
lists where each exhibit ran. Tables~\ref{tab:crossstack} and~\ref{tab:tau} and the A100
blocks of Table~\ref{tab:bands} count $\tau$ over the scored cache: a compaction fires when
the number of scored tokens reaches a multiple of $\tau$, which without the fill compacts
every few hundred decode steps. Their $\tau{>}1$ cells compare only within their own table.

\begin{table}[!htbp]
\centering
\caption{\textbf{Where each exhibit ran.} $^{a}$With torch 2.7.0 and FlashAttention-2
2.8.3. $^{b}$Its L40S blocks; its A100 blocks ran on the accuracy-grid platform. The accuracy grids ran several cells per card, so their wall-clock times are
load-confounded and are not reported; every timing comes from a run with one process
per card. Any other table states its
platform in its caption.}
\label{tab:hosts}
\small
\setlength{\tabcolsep}{4pt}
\begin{tabular}{@{}p{0.43\linewidth}llp{0.2\linewidth}@{}}
\toprule
Exhibit & GPU & Library & Load \\
\midrule
Accuracy grids and the tables built from them (Tables~\ref{tab:acc-math},
\ref{tab:termination}, \ref{tab:acc-math-family}, \ref{tab:pairs}, \ref{tab:acc-aime}),
component and interval cells (Tables~\ref{tab:components}, \ref{tab:tauq}), the loop
of Appendix~\ref{app:loop} & A100 & transformers 4.57.1$^{a}$ & two to four cells per card \\
Equal-load timing (Table~\ref{tab:time-aime}) & A100 & transformers 4.57.1$^{a}$ & one process per card \\
Kernel costs (Tables~\ref{tab:decode-step}, \ref{tab:prefill-latency}) & A100, 40\,GB & transformers 4.57.1$^{a}$ & one exclusive card \\
Prefill memory (Figure~\ref{fig:prefillmem}, right) & A100, 80\,GB & transformers 5.2.0 & one process \\
Eviction frequency and band ablations (Tables~\ref{tab:tau}, \ref{tab:bands}$^{b}$),
the fill ablation (Table~\ref{tab:fill-l40s}), the segment classifiers (Table~\ref{tab:segablation}), the right column of
Table~\ref{tab:crossstack} & L40S & transformers 5.2.0 & one process per card \\
vLLM eviction and hook cost (Tables~\ref{tab:vllm-evict}, \ref{tab:vllm}) & A100 & vLLM 0.11.2 & one engine per card \\
vLLM throughput curve (Figure~\ref{fig:vllm-tp}) & H100 & vLLM 0.11.2 & one engine \\
\bottomrule
\end{tabular}

\end{table}

A library version changes greedy traces, and with them accuracy on a fixed problem set,
by a few problems in a hundred (Table~\ref{tab:crossstack}). The two stacks
return the same hidden states up to numerics and tokenize the prompt to within a
token, so the difference is numerical, not an indexing difference. The main tables use
the stack on which our policies score lower, with every baseline measured on the same
stack, so every comparison in the paper is made within one platform.

\begin{table}[!htbp]
\centering
\caption{\textbf{The main tables use the stack on which our policies score lower.} MATH-500 accuracy (\%) on
the same 100 problems at $K{=}1024$ under two inference stacks; the left column is the
platform of Table~\ref{tab:acc-math}. The stacks return
the same hidden states up to numerics (per-layer norms within 4\% from layer 3 on,
last-token cosines 0.92--0.99) and tokenize the prompt to 29 and 28 tokens. The $\tau{=}128$ row counts $\tau$ over the scored
cache.}
\label{tab:crossstack}
\small
\setlength{\tabcolsep}{6pt}
\begin{tabular}{@{}lrr@{}}
\toprule
 & transformers 4.57.1, A100 & transformers 5.2.0, L40S \\
\midrule
\methodseg~($\tau{=}128$, scored-cache rule) & 68.0 & 74.0 \\
\methodseg~($\tau{=}1$) & 66.0 & 70.0 \\
raw score & 66.0 & 71.0 \\
\midrule
RaaS & 77.0 & 77.0 \\
H2O & 72.0 & 75.0 \\
none & 77.0 & 77.0 \\
\bottomrule
\end{tabular}

\end{table}

\FloatBarrier
\subsection{The positional convention decides whether H2O collapses}
\label{app:convention}

Where a retained token sits after eviction is a choice the harness makes
(\S\ref{sec:setup}). Under the rewound convention the survivors are re-embedded at the
compacted cache's physical slots, so their rotary positions no longer match the
distances between them in the text. H2O then collapses on reasoning traces
(Table~\ref{tab:h2o-rewound}), and RaaS reports the same collapse for H2O
\citep{raas}. Under the logical convention, which every table not labeled rewound
uses, the collapse does not occur and H2O is a competitive baseline at every budget.

\begin{table}[!htbp]
\centering
\caption{\textbf{H2O's collapse on reasoning traces comes from the positional
convention, not from the policy.} MATH-500 accuracy (\%) of H2O on Llama-8B, problems
0--99, eager attention, 8,192-token cap, with the survivors re-embedded at the
compacted cache's physical slots (rewound; another platform of this harness) and kept
at their original positions (logical; the same shard of Table~\ref{tab:acc-math}).
Cap hits are generations that reached the cap; empty are generations of no tokens.
Under rewound positions every generation at $K{=}512$ loops to the cap and most at
$K{=}1024$ stop empty; under logical positions neither failure occurs. The two
columns ran on different platforms and are not paired.}
\label{tab:h2o-rewound}
\small
\setlength{\tabcolsep}{5pt}
\begin{tabular}{@{}rrrrrrr@{}}
\toprule
 & \multicolumn{3}{c}{rewound positions} & \multicolumn{3}{c}{logical positions} \\
\cmidrule(lr){2-4}\cmidrule(lr){5-7}
$K$ & acc & cap hits & empty & acc & cap hits & empty \\
\midrule
512  & 1.0  & 100 & 0  & 61.0 & 38 & 0 \\
1024 & 5.0  & 0   & 93 & 72.0 & 21 & 0 \\
2048 & 49.0 & 0   & 48 & 78.0 & 14 & 0 \\
4096 & 67.0 & 0   & 27 & 75.0 & 19 & 0 \\
\bottomrule
\end{tabular}

\end{table}

\FloatBarrier
\subsection{What each policy always keeps}
\label{app:policies}

The policies differ in which structural tokens they exempt from scoring and in
whether a compaction fills the budget (Table~\ref{tab:structural}). H2O keeps
attention sinks and a recency window; RaaS and our policies keep the whole prompt and
a recency window; ThinKV and LongFlow keep a recency window only, and R-KV adds its
last eight observed tokens. RaaS runs its token-level algorithm \citep[Algorithm~1]{raas}. Two policies can hold
fewer than $K$ tokens after a compaction: ThinKV, whose per-segment budgets need not
sum to $K$, and \methodseg, whose tier caps bind before the budget does. The recency
window is $\min(128, K/4)$ throughout.

\begin{table}[!htbp]
\centering
\caption{\textbf{The policies keep different structural tokens, and two of them can
hold fewer than $K$.} ``Always kept'' is exempt from scoring; ``Holds'' is the cache
size after a compaction. ThinKV's segment types are reasoning, execution and
transition; R-KV keeps its last eight observed tokens.}
\label{tab:structural}
\small
\setlength{\tabcolsep}{4pt}
\setlength{\tabcolsep}{3pt}
\begin{tabular}{@{}lllc@{}}
\toprule
Method & Always kept & Ranked by & Holds \\
\midrule
H2O          & 4 sinks, recency & cumulative attention & $K$ \\
ThinKV       & recency & segment type & $\le K$ \\
RaaS         & prompt, recency & attention timestamp & $K$ \\
R-KV         & 8 observed, recency & importance minus redundancy & $K$ \\
LongFlow     & recency & attention $\times \lVert v\rVert_1$ & $K$ \\
\midrule
\methodq     & prompt, recency & tiers, query term, fill & $K$ \\
\methodseg   & prompt, recency & tiers, $s(t)$ & $\le K$ \\
\methodflat, raw score & prompt, recency & $s(t)$ or its raw form & $K$ \\
KV-key, Lag-KV & prompt, recency & key or key-and-value variance & $K$ \\
\bottomrule
\end{tabular}

\end{table}

These conventions also vary across the literature, which matters when comparing
published numbers. In LongFlow's harness, for instance, H2O receives a recency window
of half the budget and the prompt is compressed once at prefill, a larger structural
allocation than the one used here; the released evaluation code also applies a
repetition penalty to every method, which suppresses the looping that eviction induces
on reasoning traces (\S\ref{sec:r-failure}). Published H2O figures on reasoning traces
therefore depend on the structural allocation, the decoding penalty and the positional
convention (Appendix~\ref{app:convention}) as much as on the policy.

\FloatBarrier
\subsection{How R-KV and LongFlow were adapted}
\label{app:adapted}

R-KV \citep{rkv} and LongFlow \citep{longflow} are implemented in our harness from
their papers' equations and reported at $n{=}495$ on the platform of
Table~\ref{tab:acc-math}. Two adaptations make them comparable with the other rows.
First, R-KV selects tokens per head and LongFlow per layer, whereas our harness applies one keep-mask to the
whole cache; we therefore average each method's scores over heads and layers
before selection. Second, R-KV's importance and redundancy terms are standardized at
each compression step before they are mixed, since its published mixing weights
otherwise leave the redundancy term numerically inert under this averaging; its
128-step compression interval is kept. LongFlow's score, attention $\times \lVert v \rVert_1$,
is computed eagerly rather than through its released Triton kernel, which changes the
speed of the computation and not the tokens it selects. The released code documents
both adaptations.

\FloatBarrier
\subsection{Our ThinKV implementation, and a stateless variant}
\label{app:thinkvfaithful}

Every ThinKV number in this paper comes from our implementation: segment entropies
classified by a Gaussian kernel density with Silverman's robust bandwidth, boundaries
at the two deepest interior minima, and reclassification every 128 steps. We also
evaluated a simpler variant that splits the entropies at tertiles and reclassifies at
every decode step (Table~\ref{tab:thinkvfaithful}). Under the rewound-position protocol
of Appendix~\ref{app:convention}, it scores within noise of that
implementation at every budget, so in this harness neither the density fit nor the refresh
interval explains ThinKV's place. We report the
variant here and not as a baseline, because it departs further from ThinKV. The density
engages only when a cache holds enough segments for a multimodal entropy distribution,
which at 128-token segments means $K \ge 2048$; below that the implementation falls
back to tertiles, and the two variants differ only in the refresh interval.

\begin{table}[!htbp]
\centering
\caption{\textbf{A stateless variant of ThinKV scores within noise of our
ThinKV implementation.} MATH-500 accuracy (\%) on Llama-8B, $n{=}100$ per cell, whose
Wilson interval spans roughly $\pm$9\,pp, under the rewound-position protocol of
Appendix~\ref{app:convention} (not comparable to Table~\ref{tab:acc-math}). The variant splits segment entropies at
tertiles and reclassifies at every decode step; our implementation fits a
density and reclassifies every 128 steps. Below $K{=}2048$ the two differ only in the
refresh interval.}
\label{tab:thinkvfaithful}
\normalsize
\setlength{\tabcolsep}{5pt}
\begin{tabular}{@{}lrrrr@{}}
\toprule
Implementation & 512 & 1024 & 2048 & 4096 \\
\midrule
ThinKV (KDE, $\tau{=}128$) & 4.0 & 35.0 & 52.0 & 68.0 \\
Stateless tertile variant (ours)        & 6.0 & 32.0 & 58.0 & 71.0 \\
\bottomrule
\end{tabular}

\end{table}

\FloatBarrier
\section{The epiphany score}
\label{app:g-signal}

\FloatBarrier
\subsection{Counterfactual labels and the stop-token filler}
\label{app:labeldetails}

Labels come from sliding-window occlusion over the reasoning span of each correctly
answered trace. The window is 32 tokens with stride 16, so each interior position is
covered by two windows. The reasoning span ends at the \texttt{</think>} token
sequence, or failing that at the last \texttt{\textbackslash boxed\{}, or failing that
64 tokens before the end. Each window's tokens are replaced by the tokenizer's pad id,
which for this model resolves to the end-of-sequence (EOS) token, since the model
defines no pad token. The modified context up to the end of the reasoning span is fed
back, and the answer is regenerated greedily with up to 512 new tokens. A position is
labeled important if any window covering it changes the answer; prompt positions
count as important, and the answer span is not tested. Regeneration is deterministic,
so the labels are reproducible.

\paragraph{A labeled trace.} On MATH-500 problem 3 (``How many positive whole-number
divisors does 196 have?''), 3 of the 62 windows tested change the answer. One covers
the plan, the sentence that commits to the method; the other two cover the conclusion.
Every other span, the factorization included, can be removed without changing the
answer. Important tokens are marked with double brackets (Llama-8B, EOS-filler labels).

{\small
\begin{verbatim}
Okay, so I need to figure out how many positive whole-number divisors
196 has. Hmm, divisors, right? [...] I remember that to find the number
of divisors, I need to look at the prime factorization of the number.
[[ Once I have that, there's some formula or method to calculate the
number of divisors based on the exponents in the prime factors. Let me
try to]] recall how that works.

First, let me find the prime factors of 196. [...]

[...] So, yeah, 9 is[[ the right answer.

**Final Answer**
The number of positive whole-number divisors of 196 is \boxed{9}.
</think>]]
\end{verbatim}
}

\paragraph{Filler controls.} An EOS filler injects a stop signal as well as removing
content, so the labeling is repeated with three stop-free fillers on matched traces and
windows: an untrained pad token (\texttt{<|finetune\_right\_pad\_id|>}), a repeated
random in-vocabulary token, and a high-entropy filler that draws a distinct random
token per position. Paired over the same 29 traces, the three stop-free fillers agree
with one another and each differs from EOS, which attributes about a third of the EOS
flips to the stop signal (Table~\ref{tab:fillers}). Under the high-entropy filler,
across all 60 traces, every Band~A layer stays positive and every Band~B layer stays
negative, at magnitudes two to four times smaller than under EOS; the stop-signal
flips account for the difference.

\begin{table}[!htbp]
\centering
\caption{\textbf{A third of the EOS flips are stop semantics, and the anatomy holds
without them.} Flip rate per occluded window, paired over the same 29 MATH-500
traces (Wilcoxon against EOS; the three stop-free fillers agree pairwise,
$p\ge0.11$), and the mean per-layer $\rho$ over each band under the EOS labels
(Table~\ref{tab:perlayer}) and under the high-entropy filler across all 60 traces.
The two rows without band means were not relabeled at the per-layer level.}
\label{tab:fillers}
\small
\setlength{\tabcolsep}{5pt}
\begin{tabular}{@{}lrrrr@{}}
\toprule
Filler & flip rate & vs.\ EOS ($p$) & Band A mean $\rho$ & Band B mean $\rho$ \\
\midrule
EOS (pad id) & 0.176 & -- & $+0.077$ & $-0.200$ \\
untrained pad token & 0.115 & 0.023--0.052 & -- & -- \\
repeated random token & 0.114 & 0.023--0.052 & -- & -- \\
high-entropy random tokens & 0.118 & 0.023--0.052 & $+0.019$ & $-0.092$ \\
\bottomrule
\end{tabular}

\end{table}

\FloatBarrier
\subsection{Per-layer correlations}
\label{app:layers}

Table~\ref{tab:perlayer} gives the Spearman $\rho$ between $\bar{g}_l$, the
hidden-state change at layer $l$ averaged over a 64-token rolling window, and the counterfactual importance labels, for all 32
layers on the two competition-mathematics datasets under both attention back-ends; it
is the data behind Figure~\ref{fig:layers}. Band A (7--13) is positive throughout and
Band B (18--25) negative throughout on MATH-500 under both back-ends. The AIME-2024
columns keep each band's sign in the mean, at a sample size that does not support a
per-layer claim (Appendix~\ref{app:power}). The last layer flips sign across datasets and is not used.
Table~\ref{tab:attnsignal} sets the two attention-derived signals beside the deployed
band layers on the configurations where attention weights exist.

\begin{table}[!htbp]
\centering
\caption{\textbf{Band A is positive and Band B negative at every layer on MATH-500 under
both back-ends.} Spearman $\rho$ of $\bar{g}_l$ with the importance labels, all 32 layers,
Llama-8B, EOS-filler labels; Band A (7--13) and Band B (18--25) rows are in bold.}
\label{tab:perlayer}
\normalsize
\setlength{\tabcolsep}{4pt}
\begin{tabular}{@{}rrrrr@{}}
\toprule
$l$ & math500 & m500-eag & aime24 & aime24-eag \\
\midrule
0 & $-$0.173 & $-$0.199 & $-$0.208 & $+$0.037 \\
1 & $-$0.172 & $-$0.202 & $-$0.208 & $-$0.052 \\
2 & $-$0.121 & $-$0.151 & $-$0.158 & $-$0.125 \\
3 & $-$0.121 & $-$0.124 & $-$0.150 & $-$0.072 \\
4 & $-$0.153 & $-$0.139 & $-$0.191 & $-$0.052 \\
5 & $-$0.079 & $-$0.058 & $-$0.105 & $+$0.093 \\
6 & $+$0.011 & $-$0.007 & $-$0.048 & $+$0.150 \\
\textbf{7} & $+$0.079 & $+$0.047 & $-$0.010 & $+$0.156 \\
\textbf{8} & $+$0.067 & $+$0.078 & $+$0.037 & $+$0.155 \\
\textbf{9} & $+$0.082 & $+$0.107 & $+$0.054 & $+$0.130 \\
\textbf{10} & $+$0.112 & $+$0.141 & $+$0.040 & $+$0.120 \\
\textbf{11} & $+$0.092 & $+$0.144 & $+$0.079 & $+$0.120 \\
\textbf{12} & $+$0.038 & $+$0.077 & $+$0.076 & $+$0.114 \\
\textbf{13} & $+$0.071 & $+$0.089 & $+$0.009 & $+$0.065 \\
14 & $-$0.016 & $+$0.006 & $-$0.076 & $-$0.032 \\
15 & $+$0.017 & $+$0.016 & $-$0.201 & $-$0.147 \\
16 & $+$0.008 & $-$0.016 & $-$0.205 & $-$0.165 \\
17 & $-$0.003 & $-$0.005 & $-$0.173 & $-$0.191 \\
\textbf{18} & $-$0.081 & $-$0.045 & $-$0.129 & $-$0.184 \\
\textbf{19} & $-$0.147 & $-$0.074 & $-$0.099 & $-$0.208 \\
\textbf{20} & $-$0.191 & $-$0.097 & $-$0.037 & $-$0.224 \\
\textbf{21} & $-$0.209 & $-$0.109 & $-$0.011 & $-$0.227 \\
\textbf{22} & $-$0.223 & $-$0.121 & $-$0.007 & $-$0.217 \\
\textbf{23} & $-$0.254 & $-$0.151 & $+$0.007 & $-$0.200 \\
\textbf{24} & $-$0.250 & $-$0.146 & $+$0.016 & $-$0.188 \\
\textbf{25} & $-$0.243 & $-$0.142 & $-$0.034 & $-$0.187 \\
26 & $-$0.211 & $-$0.107 & $-$0.013 & $-$0.157 \\
27 & $-$0.066 & $-$0.003 & $-$0.030 & $-$0.086 \\
28 & $-$0.053 & $-$0.018 & $+$0.002 & $-$0.059 \\
29 & $+$0.082 & $+$0.049 & $-$0.042 & $+$0.030 \\
30 & $+$0.135 & $+$0.065 & $-$0.035 & $+$0.062 \\
31 & $+$0.220 & $+$0.092 & $-$0.196 & $-$0.022 \\
 
\bottomrule
\end{tabular}

\end{table}

\begin{table}[!htbp]
\centering
\caption{\textbf{Cumulative attention, the signal H2O ranks by, is a weaker correlate
of importance than either deployed band layer.} Spearman $\rho$ with the
counterfactual importance labels on the two eager configurations, where attention
weights exist: H2O's cumulative attention, attention entropy, and the
hidden-state change at the deployed layers of Band A and Band B and the KV-key variance,
each averaged over a 64-token rolling window. Attention entropy is the stronger attention-derived signal and is
competitive with the band layers, but it needs the same attention matrix.}
\label{tab:attnsignal}
\small
\setlength{\tabcolsep}{5pt}
\begin{tabular}{@{}lrrrrr@{}}
\toprule
 & \multicolumn{2}{c}{attention-derived} & \multicolumn{2}{c}{hidden-state change} & KV-key \\
\cmidrule(lr){2-3}\cmidrule(lr){4-5}
Dataset (eager) & cumulative & entropy & $\bar g_{10}$ & $\bar g_{21}$ & variance \\
\midrule
MATH-500 & $+$0.050 & $+$0.176 & $+$0.141 & $-$0.109 & $+$0.214 \\
AIME-2024 & $-$0.011 & $-$0.139 & $+$0.120 & $-$0.227 & $-$0.022 \\
 
\bottomrule
\end{tabular}

\end{table}

\FloatBarrier
\subsection{Statistical power of the correlations}
\label{app:power}

Effective sample size counts independent traces rather than token pairs, since tokens
within a trace are correlated. Table~\ref{tab:power} gives $n_{\mathrm{eff}}$ and the
resulting standard error of Spearman $\rho$ for each configuration. MATH-500 and GSM8K
have enough traces to support a per-layer claim; every AIME configuration has an
interval that spans zero. The paper therefore makes the per-layer claim on MATH-500, and
in its shifted form on GSM8K; on AIME it reports the sign of each band and makes no
per-layer claim.

\begin{table}[!htbp]
\centering
\caption{\textbf{Only MATH-500 and GSM8K have enough traces for a per-layer claim.}
Effective sample size per configuration and the standard error of Spearman $\rho$,
$1/\sqrt{n_{\mathrm{eff}}}$; $n_{\mathrm{eff}}$ counts independent traces, not scored
tokens.}
\label{tab:power}
\normalsize
\begin{tabular}{@{}lrr@{}}
\toprule
Dataset & $n_{\mathrm{eff}}$ $\uparrow$ & $\pm$SE ($\rho$) $\downarrow$ \\
\midrule
math500        & 72  & 0.118 \\
math500-eager  & 78  & 0.113 \\
gsm8k-eager    & 352 & 0.053 \\
aime2024       & 11  & 0.301 \\
aime2024-eager & 8   & 0.354 \\
aime2025/2026  & $\le 5$ & $\ge 0.45$ \\
\bottomrule
\end{tabular}

\end{table}

\FloatBarrier
\subsection{The window and the bands are not finely tuned, and Band B is dispensable}
\label{app:smoothing}
\label{app:bands}

The causal rolling window raises the correlation of every deployed signal with the
labels above its raw form, and above an exponential moving average, in every
configuration (Table~\ref{tab:smoothing}). These choices are made on $\rho$; what
matters for deployment is whether they change accuracy. Under the rewound-position
protocol of Appendix~\ref{app:convention}, moving the band layers by two
or changing the window by a factor of four leaves accuracy at $K{=}2048$ identical
across all eight variants, with the same prediction on every problem, and moves it by
one problem at $K{=}1024$, inside a sampling interval an order of magnitude wider
(Table~\ref{tab:bandw}).

\begin{table}[!htbp]
\centering
\caption{\textbf{The causal rolling window raises the correlation of every deployed
signal above its raw form.} $|\rho|$ with the counterfactual importance labels under
three smoothings, Llama-8B; the exponential moving average uses $\alpha{=}0.9$, and the
rolling window spans 64 tokens. The
AIME-2024 rows rest on few traces (Table~\ref{tab:power}). Pre-RoPE key statistics
change $|\rho|$ by at most 0.0005 in any cell and are not used.}
\label{tab:smoothing}
\small
\setlength{\tabcolsep}{5pt}
\begin{tabular}{@{}llrrr@{}}
\toprule
Configuration & Signal & raw & EMA & rolling \\
\midrule
MATH-500 & $\bar g_{10}$ (Band A) & 0.034 & 0.084 & 0.112 \\
 & $\bar g_{21}$ (Band B) & 0.085 & 0.180 & 0.209 \\
 & KV-key variance & 0.288 & 0.356 & 0.380 \\
\addlinespace
MATH-500 (eager) & $\bar g_{10}$ (Band A) & 0.041 & 0.101 & 0.141 \\
 & $\bar g_{21}$ (Band B) & 0.043 & 0.089 & 0.109 \\
 & KV-key variance & 0.150 & 0.193 & 0.214 \\
\addlinespace
AIME-2024 & $\bar g_{10}$ (Band A) & 0.003 & 0.014 & 0.040 \\
 & $\bar g_{21}$ (Band B) & 0.010 & 0.010 & 0.011 \\
 & KV-key variance & 0.203 & 0.246 & 0.259 \\
\addlinespace
AIME-2024 (eager) & $\bar g_{10}$ (Band A) & 0.037 & 0.094 & 0.120 \\
 & $\bar g_{21}$ (Band B) & 0.067 & 0.173 & 0.227 \\
 & KV-key variance & 0.014 & 0.018 & 0.022 \\
 
\bottomrule
\end{tabular}

\end{table}

\begin{table}[!htbp]
\centering
\caption{\textbf{The bands and the window are not fitted parameters at this
granularity.} MATH-500 accuracy (\%) with the band layers $(a,b)$ and rolling window
$W$ perturbed around the deployed $(10,21,64)$, $n{=}100$ per cell, measured on the
global top-$K$ form (\methodflat) under the rewound-position protocol of Appendix~\ref{app:convention}
(Llama-8B, FlashAttention-2; not comparable to Table~\ref{tab:acc-math}). At
$K{=}2048$ every variant returns the same prediction on all 100 problems. At
$K{=}1024$ the eight span 21--22\%, a one-problem range inside a Wilson
interval of roughly $\pm$9\,pp.}
\label{tab:bandw}
\normalsize
\setlength{\tabcolsep}{5pt}
\begin{tabular}{@{}llrr@{}}
\toprule
Variant & $(a,b,W)$ & $K{=}1024$ & $K{=}2048$ \\
\midrule
deployed & 10, 21, 64 & 21.0 & 49.0 \\
\midrule
A$-$2 & 8, 21, 64 & 21.0 & 49.0 \\
A$+$2 & 12, 21, 64 & 22.0 & 49.0 \\
B$-$2 & 10, 19, 64 & 22.0 & 49.0 \\
B$+$2 & 10, 23, 64 & 21.0 & 49.0 \\
both $-$2 & 8, 19, 64 & 21.0 & 49.0 \\
both $+$2 & 12, 23, 64 & 21.0 & 49.0 \\
window 32 & 10, 21, 32 & 22.0 & 49.0 \\
window 128 & 10, 21, 128 & 21.0 & 49.0 \\
\bottomrule
\end{tabular}

\end{table}

The score is the difference of two band $z$-scores, $z_{10}-z_{21}$
(\S\ref{sec:signal}). Table~\ref{tab:bands} asks what each band contributes, first on
\methodflat, a global top-$K$ over the detrended score with no segment
tiering, so that nothing but the score changes between rows, and then on
\methodseg~itself on two platforms and two models.
Every form is within paired noise of the deployed difference: on \methodflat, Band A
alone (14 wins / 6 losses), Band B alone and the two alternative layer pairs; on
\methodseg, Band A alone in every cell on both platforms and both models. The deployed
$z_{10}-z_{21}$, the form the correlations of \S\ref{sec:signal} motivate, is therefore kept.

\begin{table}[!htbp]
\centering
\caption{\textbf{Subtracting Band B, dropping it, and moving the bands all change accuracy
within paired noise.} MATH-500 accuracy (\%) on problems 0--99, greedy; in parentheses,
wins/losses against the A$-$B row of the same block. L40S: transformers 5.2.0; A100: the
platform of Table~\ref{tab:acc-math}, with $\tau$ counted over the scored cache (Appendix~\ref{app:impl}). Rows compare only within a block.}
\label{tab:bands}
\small
\setlength{\tabcolsep}{4pt}
\begin{tabular}{@{}llrrrr@{}}
\toprule
Setting & Score & $K{=}512$ & $K{=}1024$ & $K{=}2048$ & $K{=}4096$ \\
\midrule
\multicolumn{6}{@{}l}{\emph{\methodflat~(global top-$K$), Llama-8B, L40S, $\tau{=}1$}}\\
& A$-$B, layers 10/21 (deployed) & -- & 62 & 71 & -- \\
& A only (layer 10) & -- & 70 (14/6) & 74 (4/1) & -- \\
& B only (layer 21) & -- & 63 (5/4) & 74 (5/2) & -- \\
& difference, layers 7/12 & -- & 64 (7/5) & 72 (4/3) & -- \\
& difference, layers 5/15 & -- & 59 (7/10) & 71 (3/3) & -- \\
\midrule
\multicolumn{6}{@{}l}{\emph{\methodseg, Llama-8B, L40S, $\tau{=}1$}}\\
& A$-$B & -- & 70 & 77 & -- \\
& A only & -- & 74 (8/4) & 75 (3/5) & -- \\
\midrule
\multicolumn{6}{@{}l}{\emph{\methodseg, Llama-8B, A100, $\tau{=}128$ (scored-cache rule)}}\\
& A$-$B & 65 & 68 & 76 & 73 \\
& A only & 64 (5/6) & 69 (6/5) & 69 (4/11) & 76 (4/1) \\
\midrule
\multicolumn{6}{@{}l}{\emph{\methodseg, Qwen-7B, A100, $\tau{=}128$ (scored-cache rule; layers 20/24)}}\\
& A$-$B & 67 & 74 & 82 & 83 \\
& A only & 65 (5/7) & 75 (8/7) & 80 (2/4) & 84 (2/1) \\
\bottomrule
\end{tabular}

\end{table}

\FloatBarrier
\subsection{The two-band anatomy on Qwen-7B}
\label{app:qwen}

Every layer index above was located on Llama-8B. We ran the same procedure on
Qwen-7B (28 layers): eager traces on MATH-500, counterfactual
labels, and the per-layer Spearman $\rho$ of the hidden-state change, averaged over a
64-token rolling window, against importance, over 123,599 labeled positions. The shape replicates. A positive band
spans layers 14--22 and peaks at layer 20 ($\rho{=}{+}0.122$), with a negative band
directly above it at layers 24--27, strongest at layer 24 ($-0.144$). Taking the most
positive layer of the lower-mid network and the most negative layer of the upper-mid
network gives layers 20 and 24, the score layers of every Qwen-7B result in the paper.

Two things differ from Llama-8B. Qwen-7B's strongest negative correlations sit at layers
0--2 ($-0.27$ to $-0.30$), twice the magnitude of layer 24, and Llama-8B has no such
early band; restricting the choice to the mid network excludes them, so the shared
anatomy is the mid-network shape, not the location of the strongest signal. The Qwen
labels also use the EOS filler only. Neither point affects the accuracy results, which
take the two layers as given. The same procedure finds the same shape on a second model; the layer indices are
per model.

\FloatBarrier
\subsection{On GSM8K the bands move}
\label{app:gsm8k}

On GSM8K, labeled on 355 correctly answered traces ($n_{\mathrm{eff}}{=}352$), the
anatomy shifts (Table~\ref{tab:gsm8k}). The positive signal moves to the early layers,
the negative band extends across most of the network, and the last layer turns strongly
positive; attention entropy and the KV-key variance both reverse sign relative to
MATH-500. On this reading, harder problems carry the load-bearing computation
in the mid layers and simpler arithmetic in the early layers. GSM8K traces are also much
shorter, however, and are labeled with the EOS filler only, so a trace-length effect is
not separated from a difficulty effect. GSM8K therefore serves as a probe of how the
anatomy depends on the task, not as an accuracy benchmark.

\begin{table}[!htbp]
\centering
\caption{\textbf{On GSM8K the positive band moves to the early layers, the negative
band widens, and two signals reverse sign.} Spearman $\rho$ with the counterfactual
importance labels, Llama-8B, eager attention, for the deployed band layers, three
other layers, attention entropy and the KV-key variance, each averaged over a 64-token
rolling window. GSM8K is labeled with the EOS filler only.}
\label{tab:gsm8k}
\small
\setlength{\tabcolsep}{6pt}
\begin{tabular}{@{}lrr@{}}
\toprule
Signal & MATH-500 (eager) & GSM8K (eager) \\
\midrule
$\bar g_{0}$ & $-$0.199 & $+$0.181 \\
$\bar g_{7}$ & $+$0.047 & $+$0.076 \\
$\bar g_{10}$ (Band A, deployed) & $+$0.141 & $-$0.048 \\
$\bar g_{15}$ & $+$0.016 & $-$0.351 \\
$\bar g_{21}$ (Band B, deployed) & $-$0.109 & $-$0.303 \\
$\bar g_{31}$ & $+$0.092 & $+$0.231 \\
attention entropy & $+$0.176 & $-$0.313 \\
KV-key variance & $+$0.214 & $-$0.261 \\
\bottomrule
\end{tabular}

\end{table}

\FloatBarrier
\section{The policy}
\label{app:g-policy}

\FloatBarrier
\subsection{What the policy keeps}
\label{app:retained}

The name ``epiphany score'' claims that high-scoring tokens are the substantive ones:
intermediate results, committed steps, the answer forming. Figure~\ref{fig:prefillmem} (left)
shows what the score alone keeps on a real trace; here we show the shipped policy on the
same problem. \methodq~at $K{=}1024$, in the
configuration of Table~\ref{tab:acc-math} ($\tau{=}128$, logical positions), on
MATH-500 problem 5 (``A regular hexagon can be divided into six equilateral triangles.
If the perimeter of one of the triangles is 21 inches, what is the perimeter of the
regular hexagon?''), greedy, Llama-8B: 1,588 tokens generated after a 61-token prompt,
five evictions (at tokens 963, 1,091, 1,219, 1,347 and 1,475), and 1,075 tokens retained
at the end, 834 of the 1,347 that were ranked and 241 that sat in the recency window of
the last eviction or were generated after it and were never ranked. The trace
terminated with the correct answer, 42. Retained tokens are bracketed and evicted tokens
are not; the never-ranked tail is omitted.

{\small
\begin{verbatim}
[Okay][,][ so][ I][ have] this[ problem] here: A[ regular][ hex][agon]
[ can][ be][ divided][ into][ six][ equ][ilateral][ triangles][.] [...]
[ Hmm][,][ let][ me] try to figure[ this][ out][ step][ by][ step][.]
[...] It's also[ mentioned] that it can be divided into[ six] equ
[ilateral][ triangles][.] [...] if you draw lines from the[ center] of
[ the] hexagon to each of its vertices, you'll end up with six [...]
[Let][ me][ denote] the[ length] of one[ side] of[ the][ equ]ilateral
[ triangle][ as][ '][s]['.] [...] [3][s][ =][ ][21] [...] [s][ =][ ][7]
[ inches] [...]
[I][ remember][ that] in a regular[ hex][agon][,] the radius[ (]distance
from center to a vertex) is[ equal] to the[ length] of the side of the
hex[agon][.] Is that[ right][?][ Let] me recall.[ Yes], in a regular
hexagon, the radius of the circumscribed circle is equal to the length
of[ each] side[.][ So][,][ the] radius[ is] equal to[ '][s][',][ which]
[ is][ ][7][ inches][.] [...] Per[imeter] of hex[agon][ =] [6][ *][ s]
[ =][ ][6][ *][ ][7][ =][ ][42][ inches][.]
[Wait],[ but][ hold][ on][,][ let][ me][ make] sure[ I][ didn]'t make
[ a][ mistake][.] [...] [then][ ][2][r][ +][ s][ =][ ][21][.][ Since]
[ r][ =][ s][,][ that][ gives] [2][s][ +][ s][ =][ ][3][s][ =][ ][21]
[,][ so][ s][ =][ ][7][.] [...] [so][ the][ perimeter][ is][ ][6][*][7]
[=][42][ inches][.] That seems[ correct][.]
\end{verbatim}
}

\begin{figure}[p]
\centering
\includegraphics[width=\textwidth,height=0.8\textheight,keepaspectratio]{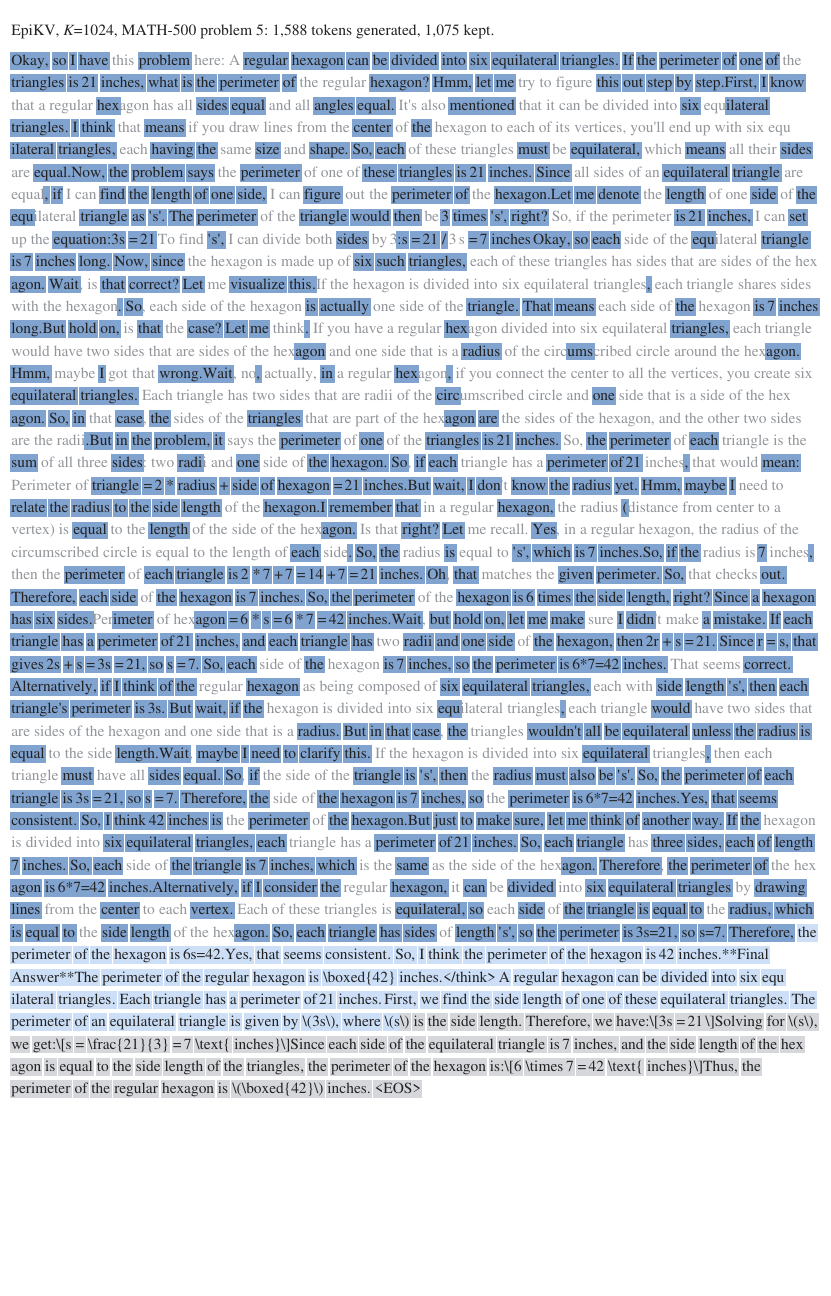}
\caption{\textbf{What survives is the problem's objects and the steps that carry
the argument; what goes is the connective prose.} Every generated token of the trace
above in reading order: the problem of Figure~\ref{fig:prefillmem} (left), under
\methodq~rather than \methodseg. Blue-highlighted tokens survived eviction by relevance, pale-blue ones were never ranked against a full cache, grey-highlighted ones sat in the recency window, and plain grey text was evicted. Of the 1,347 ranked tokens 834 survive, 228 of the first
400 and 288 of the last 400. The cache held at most 1,151 entries against a budget of
1,024, the overshoot the boundary rule allows (Appendix~\ref{app:tau}).}
\label{fig:retainedfull}
\end{figure}

The policy keeps the problem's objects and the plan (``regular hexagon'', ``divided into
six equilateral triangles'', ``step by step'', ``denote the length of one side \ldots
as `s'\,''), and, where the trace turns, the turn itself: ``the radius is equal to `s',
which is 7 inches'', ``2r + s = 21. Since r = s, that gives 2s + s = 3s = 21, so s =
7'' and ``6*7=42 inches'' are retained word for word. It drops the discourse around
them: ``this problem here'', ``try to figure'', ``if you draw lines from the
center of the hexagon to each of its vertices'', ``Let me recall'', a restated
definition of the circumradius, and most function words. Early tokens survive at a
high rate as well as late ones (Figure~\ref{fig:retainedfull}), so recency does not
explain the selection. Because \methodq~spends the whole budget, it keeps most of
the ranked tokens here; \methodseg~on its own trace of this problem keeps far fewer and
evicts the connective prose more aggressively (Figure~\ref{fig:prefillmem}, left). The
marked traces are released with the code, including problems 1 and 4 at the same
setting, the \methodseg~traces, and an H2O trace of problem 1.

\FloatBarrier
\subsection{Attention does not make a measurably better segment classifier}
\label{app:segablation}

The within-segment ranker is fixed to the hidden-state score, and only the signal that
classifies segments into tiers varies (Table~\ref{tab:segablation}). \methodq~and \methodseg~use key variance. The entropy of key variance is the closer
attention-free analogue of ThinKV's criterion, since low entropy marks concentrated,
reasoning-like segments. Attention sparsity is ThinKV's own criterion; it needs the
attention matrix, so it bounds what an attention-free classifier gives up. A fourth row
folds attention into the token score instead, scoring each token by its cumulative
attention times its hidden-state score with the segments unchanged.

No classifier separates from key variance at either budget (Table~\ref{tab:segablation}).
Attention sparsity trails it by one to two points, as does the entropy
variant; attention folded into the token score is two points
below it at $K{=}1024$ and four above at $K{=}2048$, and no pair is significant. An
attention-free classifier therefore gives up nothing measurable.

\begin{table}[!htbp]
\centering
\caption{\textbf{No segment classifier separates from key variance, with or without
attention.} MATH-500 accuracy (\%) with the same within-segment hidden-state ranker and
three segment classifiers, and with attention folded into the token score; Llama-8B,
problems 0--99, greedy, eager attention, eviction at every step, logical positions, L40S
(transformers 5.2.0). In parentheses, wins/losses against key variance on the same problems;
no pair is significant (exact McNemar, every $p \ge 0.29$). Not comparable to
Table~\ref{tab:acc-math}.}
\label{tab:segablation}
\normalsize
\setlength{\tabcolsep}{5pt}
\begin{tabular}{@{}llll@{}}
\toprule
Segment classifier & FA2 & $K{=}1024$ & $K{=}2048$ \\
\midrule
Key variance (\methodseg)     & \cmark & 73.0 & 75.0 \\
Entropy of key variance      & \cmark & 71.0 (5/7) & 74.0 (5/6) \\
Attention sparsity (ThinKV's) & \xmark & 72.0 (5/6) & 73.0 (3/5) \\
\midrule
Attention folded into the token score & \xmark & 71.0 (3/5) & 79.0 (6/2) \\
\bottomrule
\end{tabular}
\end{table}

\FloatBarrier
\subsection{The query term and the fill help only together}
\label{app:budgetfill}

\methodseg's tier caps (64, 32 or 8 tokens per 128-slot segment) can bind before the
budget does, so the policy can hold fewer tokens than it may; on long AIME traces at
$K{=}8192$ it holds under half of the budget. One repair spends the leftover budget on the highest-scoring evicted tokens by the
same frozen score. Under the rewound-position protocol of Appendix~\ref{app:convention},
Table~\ref{tab:fill-l40s} compares that fill with the unfilled policy on the same 100
problems. The fill is worse at every budget, significantly at the two
that bind: it solves no problem at $K{=}512$ or 1024 that the unfilled policy misses,
and the cap-hit rate rises wherever it adds tokens. More tokens chosen by a frozen score
mean fewer generations that stop, in the direction \S\ref{sec:r-failure} predicts. On
the pooled AIME set at $K{=}8192$, where the budget does not bind, the same fill is
neutral, within one problem.

\begin{table}[!htbp]
\centering
\caption{\textbf{The same test under the rewound-position protocol.}
MATH-500 accuracy (\%) and cap hits on problems 0--99, Llama-8B, \methodseg~without and
with the fill; paired wins/losses of the filled variant against the unfilled one, with
the exact McNemar $p$. L40S, transformers 5.2.0, under the rewound-position
protocol of Appendix~\ref{app:convention}; not comparable to Table~\ref{tab:acc-math}.}
\label{tab:fill-l40s}
\small
\begin{tabular}{@{}lrrrrl@{}}
\toprule
$K$ & \methodseg & cap-hit & $+$fill & cap-hit & paired (wins/losses, $p$) \\
\midrule
512  & 8.0  & 95 & 1.0  & 100 & 0/7, $0.016$ \\
1024 & 39.0 & 60 & 24.0 & 90  & 0/15, $0.0001$ \\
2048 & 62.0 & 32 & 57.0 & 49  & 1/6, $0.125$ \\
4096 & 71.0 & 24 & 70.0 & 26  & 1/2, $1.0$ \\
\bottomrule
\end{tabular}

\end{table}

The same holds under the paper's protocol. On the platform of Table~\ref{tab:acc-math}
and under the compaction rule of its shipped rows, the fill alone raises the cap-hit count in three of the four cells (from
38 to 43 and from 23 to 38 on Llama-8B at $K{=}512$ and 1024, and from 16 to 24 on
Qwen-7B at $K{=}1024$; unchanged at 35 on Qwen-7B at $K{=}512$), costs accuracy in three
and helps in none significantly, and the query term alone stays within paired noise of
\methodseg~in every cell (Table~\ref{tab:components}). Together they separate from
\methodseg, significantly in two of the four cells. The filled tokens are the
highest-scoring evicted tokens by a score frozen when they were generated, which is
the set the model has stopped reading; ranked by the query term instead, the same budget
helps. \methodq~therefore ships the filled cache ranked by relevance, and
\methodseg~stays in every table as the ablation.

\begin{table}[t]
\centering
\caption{\textbf{Neither the query term nor the fill helps reliably on its own;
together they make up \methodq.} MATH-500 accuracy (\%) on problems 0--99 with the
components of \methodq~added to the query-blind \methodseg~one at a time, on the
platform of Table~\ref{tab:acc-math}; in parentheses, paired wins/losses against
\methodseg~on the same problems, with exact McNemar stars ($^{*}$ $p<0.05$).
``--'' marks a cell that was not run. $w$ is the query window in tokens. Every row
compacts every $\tau{=}128$ decode steps, the rule of the shipped rows of
Table~\ref{tab:acc-math}, whose first shard these cells share.}
\label{tab:components}
\small
\setlength{\tabcolsep}{4pt}
\begin{tabular}{@{}lllll@{}}
\toprule
 & \multicolumn{2}{c}{Llama-8B} & \multicolumn{2}{c}{Qwen-7B} \\
\cmidrule(lr){2-3}\cmidrule(lr){4-5}
Variant & $K{=}512$ & $K{=}1024$ & $K{=}512$ & $K{=}1024$ \\
\midrule
\methodseg~(no query term, no fill) & 59.0 & 66.0 & 65.0 & 78.0 \\
+ fill by $s(t)$, no query term & 63.0 (8/4) & 62.0 (6/10) & 61.0 (4/8) & 73.0 (1/6) \\
+ query term ($w{=}8$), no fill & 60.0 (6/5) & 71.0 (10/5) & 65.0 (6/6) & 76.0 (2/4) \\
+ query term ($w{=}1$), no fill & -- & 69.0 (7/4) & -- & -- \\
+ query term ($w{=}1$) + fill & -- & 76.0 (15/5)$^{*}$ & -- & 78.0 (1/1) \\
\rowcolor{epirow}\methodq~($w{=}8$ + fill) & 66.0 (11/4) & 76.0 (12/2)$^{*}$ & 73.0 (9/1)$^{*}$ & 79.0 (5/4) \\
 
\bottomrule
\end{tabular}

\end{table}

\FloatBarrier
\subsection{Compacting less often matches compacting every step}
\label{app:tau}

\methodseg~scores every token as it is generated but compacts the cache only at
segment boundaries (\S\ref{sec:policies}). Table~\ref{tab:tau} varies how often it
compacts, from every step ($\tau{=}1$) to $\tau{=}32$ and $\tau{=}128$ counted over the
scored cache (Appendix~\ref{app:impl}), on the same 100 problems, one process per card, beside two controls: no eviction, and a score-only run
($\tau{=}\infty$: the score is computed at every step and the cache is never
compacted), which separates the cost of the score from the cost of eviction.
Accuracy is within paired noise at every budget; the largest paired difference, 6 wins / 2
losses at $K{=}1024$, is not significant. The decode rate is flat within 7\% across the rows,
and the score itself costs nothing measurable, since the score-only run reproduces the
no-eviction rate and accuracy. Wall time per problem follows the number of tokens generated,
which follows termination.

\begin{table}[!htbp]
\centering
\caption{\textbf{Compacting less often matches compacting every step.} \methodseg~on
Llama-8B, MATH-500 problems 0--99, greedy, 8,192-token cap, one process per L40S card
(transformers 5.2.0). W/L: problems solved by the row and not by $\tau{=}1$, and the reverse;
the no-eviction row covers problems 25--99 ($n{=}75$); $\tau{>}1$ is counted over the scored cache
(Appendix~\ref{app:impl}). Not comparable to Table~\ref{tab:acc-math}.}
\label{tab:tau}
\small
\setlength{\tabcolsep}{4pt}
\begin{tabular}{@{}rlrrrrrr@{}}
\toprule
$K$ & Policy & Acc.\ & W/L vs.\ $\tau{=}1$ & Cap-hit \% & Tokens & Tok/s & s/problem \\
\midrule
512 & no eviction & 77 & 10/2 & 13 & 2,974 & 31.9 & 93 \\
512 & $\tau{=}1$ (every step) & 64 & -- & 37 & 4,062 & 31.7 & 128 \\
512 & $\tau{=}32$ & 67 & 5/2 & 34 & 3,927 & 30.5 & 129 \\
512 & $\tau{=}128$ & 63 & 3/4 & 36 & 4,023 & 29.8 & 135 \\
\midrule
1024 & no eviction & 77 & 5/2 & 13 & 2,974 & 31.9 & 93 \\
1024 & score only ($\tau{=}\infty$) & 77 & 9/2 & 14 & 3,056 & 31.8 & 96 \\
1024 & $\tau{=}1$ (every step) & 70 & -- & 25 & 3,564 & 31.0 & 115 \\
1024 & $\tau{=}32$ & 69 & 3/4 & 26 & 3,656 & 30.4 & 120 \\
1024 & $\tau{=}128$ & 74 & 6/2 & 19 & 3,206 & 29.8 & 108 \\
\midrule
2048 & no eviction & 77 & 3/1 & 13 & 2,974 & 31.9 & 93 \\
2048 & score only ($\tau{=}\infty$) & 77 & 3/3 & 14 & 3,056 & 31.9 & 96 \\
2048 & $\tau{=}1$ (every step) & 77 & -- & 15 & 3,031 & 31.0 & 98 \\
2048 & $\tau{=}32$ & 77 & 3/3 & 15 & 3,053 & 30.5 & 100 \\
2048 & $\tau{=}128$ & 76 & 3/4 & 16 & 3,103 & 29.7 & 104 \\
\midrule
4096 & no eviction & 77 & 1/2 & 13 & 2,974 & 31.9 & 93 \\
4096 & $\tau{=}1$ (every step) & 80 & -- & 14 & 3,098 & 31.2 & 99 \\
4096 & $\tau{=}32$ & 80 & 2/2 & 13 & 3,034 & 30.5 & 99 \\
4096 & $\tau{=}128$ & 80 & 3/3 & 13 & 3,019 & 29.8 & 101 \\
\bottomrule
\end{tabular}

\end{table}

For \methodq, compacting more often is never better (Table~\ref{tab:tauq}). The three
short intervals score alike, as do the three long ones, and the split between them is
within paired noise at $n{=}100$ but consistent in direction. Both policies ship the
amortized schedule, which is also the form a paged engine can run
(\S\ref{sec:policies}).

\begin{table}[!htbp]
\centering
\caption{\textbf{Compacting more often does not help \methodq.} MATH-500 accuracy
(\%) on problems 0--99, Llama-8B, $K{=}1024$, on the platform of
Table~\ref{tab:acc-math}, against the compaction interval.}
\label{tab:tauq}
\small
\setlength{\tabcolsep}{6pt}
\begin{tabular}{@{}lrrrrrr@{}}
\toprule
$\tau$ (decode steps) & 1 & 8 & 16 & 32 & 64 & 128 \\
\midrule
\methodq~accuracy & 70 & 70 & 71 & 77 & 76 & 76 \\
\bottomrule
\end{tabular}

\end{table}

Two mechanical facts follow from the boundary rule and are visible on the released
traces (Appendix~\ref{app:retained}). A compaction fires only when the cache is over
budget at a boundary, so between boundaries the cache holds up to $K{+}\tau{-}1$
entries. An \methodseg~compaction lands below $K$ because the tier caps bind
(Appendix~\ref{app:budgetfill}), whereas an \methodq~compaction lands at $K$, so the
occupancy saw-tooths around $K$; the counts for the trace of
Figure~\ref{fig:retainedfull} are in its caption. The overshoot is a fraction of $K$ that
shrinks as $K$ grows, against one entry for the per-step form. A strict variant that fires when the cache exceeds $K{-}\tau{+}1$ would
remove the overshoot; we have not run it.

\FloatBarrier
\section{Full accuracy results}
\label{app:fullresults}

This section gives the tables behind \S\ref{sec:r-acc}: the members of the
\methodname~family on the first MATH-500 shard (Table~\ref{tab:acc-math-family}), the
paired counts behind the daggers of Table~\ref{tab:acc-math} (Table~\ref{tab:pairs}),
pooled AIME (Table~\ref{tab:acc-aime}) and GPQA-Diamond (Table~\ref{tab:gpqa}). AIME pools the
2024, 2025 and 2026 sets, $n{=}90$; both it and GPQA-Diamond ($n{=}198$) run with a 16,384-token
cap, so that the uncapped model's own non-termination is measured rather than imposed. \cmark~marks the policies that run on FlashAttention.

\paragraph{The family.} At $n{=}495$ (Table~\ref{tab:acc-math}) the query term and the fill
are worth $+8.7$, $+7.1$, $+2.6$ and $-0.8$ points on Llama-8B and $+3.1$, $+4.5$, $-0.2$ and
$-1.6$ on Qwen-7B over \methodseg, from the tightest budget to the loosest.
Table~\ref{tab:acc-math-family} shows where the rest of the policy earns its place.
Segment-tiered allocation carries the tight budgets: \methodseg~leads \methodflat~by 8 and 9
points on Llama-8B and by 11 and 7 on Qwen-7B at $K{=}512$ and 1024. On Qwen-7B the $z$-scored
score leads the raw magnitude only once it is allocated by segment, since \methodflat~trails
the raw score at $K{=}512$ and 2048 and \methodseg~leads it at every budget; on Llama-8B \methodflat~trails and \methodseg~is level.

\begin{table}[!htbp]
\centering
\caption{\textbf{Removing the query term or the segments costs accuracy at the tight
budgets.} MATH-500 accuracy (\%) on problems 0--99, the first of the five shards behind
Table~\ref{tab:acc-math}; the \methodq, \methodseg~and Lag-KV rows are that shard of its runs,
the others were run on this shard only. Rows below \methodq~remove its components in turn;
KV-key and Lag-KV score from the cached KV vectors. Bold: best in a column.}
\label{tab:acc-math-family}
\small
\setlength{\tabcolsep}{3.5pt}
\begin{tabular}{@{}llrrrrrrrr@{}}
\toprule
 & & \multicolumn{4}{c}{Llama-8B} & \multicolumn{4}{c}{Qwen-7B} \\
\cmidrule(lr){3-6}\cmidrule(lr){7-10}
Method & FA2 & 512 & 1024 & 2048 & 4096 & 512 & 1024 & 2048 & 4096 \\
\midrule
\rowcolor{epirow}\methodq & \cmark & \textbf{66.0} & \textbf{76.0} & \textbf{76.0} & \textbf{75.0} & \textbf{73.0} & \textbf{79.0} & 80.0 & 81.0 \\
\methodseg~(no query term) & \cmark & 59.0 & 66.0 & 73.0 & 71.0 & 65.0 & 78.0 & 80.0 & 82.0 \\
\methodflat~(no segments either) & \cmark & 51.0 & 57.0 & 66.0 & 73.0 & 54.0 & 71.0 & 74.0 & \textbf{83.0} \\
raw score (no detrending) & \cmark & 58.0 & 66.0 & 72.0 & 74.0 & 57.0 & 70.0 & 76.0 & 81.0 \\
KV-key & \cmark & 58.0 & 66.0 & 69.0 & 72.0 & 69.0 & 77.0 & \textbf{82.0} & 80.0 \\
Lag-KV & \cmark & 54.0 & 65.0 & 69.0 & 73.0 & 40.0 & 57.0 & 75.0 & 81.0 \\
\bottomrule
\end{tabular}

\end{table}

\begin{table}[!htbp]
\centering
\caption{\textbf{Where \methodq~separates from a row and where it does not.} Problems
won / lost by \methodq~against each row of Table~\ref{tab:acc-math} on the same
$n{=}495$ problems; the stars mark exact McNemar significance ($^{*}$ $p<0.05$,
$^{**}$ $p<0.01$, $^{***}$ $p<0.001$). Against RaaS no Llama-8B budget separates the
two policies and every Qwen-7B budget from 1024 upward does; against the query-blind
\methodseg~the separation sits at the tight budgets.}
\label{tab:pairs}
\small
\setlength{\tabcolsep}{3pt}
\begin{tabular}{@{}lrrrrrrrr@{}}
\toprule
 & \multicolumn{4}{c}{Llama-8B} & \multicolumn{4}{c}{Qwen-7B} \\
\cmidrule(lr){2-5}\cmidrule(lr){6-9}
\methodq~vs. & 512 & 1024 & 2048 & 4096 & 512 & 1024 & 2048 & 4096 \\
\midrule
none (no eviction) & 32/49 & 28/26 & 23/15 & 9/6 & 10/74$^{***}$ & 14/33$^{**}$ & 10/15 & 7/11 \\
\methodseg & 62/19$^{***}$ & 54/19$^{***}$ & 38/25 & 17/21 & 41/26 & 38/16$^{**}$ & 20/21 & 12/20 \\
Lag-KV & 80/12$^{***}$ & 66/17$^{***}$ & 37/14$^{**}$ & 10/6 & 162/6$^{***}$ & 110/3$^{***}$ & 41/6$^{***}$ & 11/8 \\
R-KV & 113/7$^{***}$ & 64/14$^{***}$ & 40/35 & 34/44 & 184/6$^{***}$ & 85/10$^{***}$ & 51/22$^{***}$ & 46/25$^{*}$ \\
ThinKV & 204/8$^{***}$ & 163/9$^{***}$ & 110/17$^{***}$ & 59/34$^{*}$ & 251/6$^{***}$ & 227/6$^{***}$ & 145/9$^{***}$ & 62/14$^{***}$ \\
RaaS & 30/22 & 35/33 & 38/40 & 34/40 & 39/35 & 50/24$^{**}$ & 48/24$^{**}$ & 48/20$^{***}$ \\
LongFlow & 122/11$^{***}$ & 64/25$^{***}$ & 39/32 & 35/43 & 174/11$^{***}$ & 99/16$^{***}$ & 55/24$^{***}$ & 47/21$^{**}$ \\
H2O & 67/18$^{***}$ & 48/30 & 34/36 & 34/43 & 92/25$^{***}$ & 67/18$^{***}$ & 58/21$^{***}$ & 54/20$^{***}$ \\
 
\bottomrule
\end{tabular}

\end{table}

\paragraph{Pooled AIME.} No row separates from \methodseg~at either budget on either model
once the 40 paired comparisons in Table~\ref{tab:acc-aime} are corrected (Bonferroni). The
smallest uncorrected $p$ is 0.02, for raw score and KV-key over \methodseg~at Llama
$K{=}8192$ (0 wins / 7 losses) and for \methodseg~over H2O at Qwen $K{=}8192$ (15/4).
\methodseg~is level with no eviction (Llama 8192: 0/4, $p{=}0.12$; Qwen 8192: 13/7,
$p{=}0.26$), and \methodq~is level with \methodseg~on Llama-8B at 4096 (4/3) and ahead at 8192 (5/0, $p{=}0.06$),
and slightly below it on Qwen-7B, within noise (5/9, $p{=}0.42$; 8/14, $p{=}0.29$). A single cell's
Wilson 95\% interval spans about $\pm$9 points, so the table bounds what any policy can
recover at this trace length.

\begin{table}[!htbp]
\centering
\caption{\textbf{On pooled AIME no policy separates from another after correction.} Accuracy (\%) against the cache budget;
AIME 2024--2026 pooled, $n{=}90$, greedy, 16,384-token cap. $^{\star}$No eviction is
budget-independent; it fails to terminate on 61\% (Llama-8B) and 62\% (Qwen-7B) of these
problems. No cell is marked best (paired tests in the text).}
\label{tab:acc-aime}
\normalsize
\setlength{\tabcolsep}{3.5pt}
\begin{tabular}{@{}llrrrr@{}}
\toprule
 & & \multicolumn{2}{c}{Llama-8B} & \multicolumn{2}{c}{Qwen-7B} \\
\cmidrule(lr){3-4}\cmidrule(lr){5-6}
Method & FA2 & 4096 & 8192 & 4096 & 8192 \\
\midrule
none$^{\star}$ & \cmark & 26.7 & 26.7 & 32.2 & 32.2 \\
\midrule
\rowcolor{epirow}\methodq & \cmark & 25.6 & 27.8 & 27.8 & 32.2 \\
\methodseg~($\tau{=}128$) & \cmark & 24.4 & 22.2 & 32.2 & 38.9 \\
\methodflat & \cmark & 25.6 & 25.6 & 31.1 & 37.8 \\
raw score & \cmark & 30.0 & 30.0 & 26.7 & 35.6 \\
KV-key & \cmark & 27.8 & 30.0 & 31.1 & 38.9 \\
Lag-KV & \cmark & 26.7 & 28.9 & 27.8 & 41.1 \\
\midrule
RaaS & \xmark & 23.3 & 25.6 & 27.8 & 30.0 \\
H2O & \xmark & 26.7 & 26.7 & 25.6 & 26.7 \\
R-KV & \xmark & 26.7 & 25.6 & 26.7 & 28.9 \\
LongFlow & \xmark & 28.9 & 25.6 & 26.7 & 28.9 \\
ThinKV & \xmark & 14.4 & 23.3 & 20.0 & 23.3 \\
\bottomrule
\end{tabular}

\end{table}

\FloatBarrier
\paragraph{GPQA-Diamond.} The 198 questions are four-way multiple choice; the answer is a
letter in the boxed span, and the prompt is the harness's MATH prompt with the lettered
options in the question. The uncapped model reaches the cap on 76 problems on Llama-8B and 94
on Qwen-7B, and raising the cap to 32,768 on two shards turns none of the twelve cap hits
there into an answer (eleven run to the longer cap, one stops wrong). \methodq~is within
paired noise of the uncapped model at every budget on both models, the largest paired
difference being 15 wins / 20 losses (Llama-8B, $K{=}1024$, $p{=}0.50$), and within noise of
RaaS. On Llama-8B it separates from \methodseg~at the two tight budgets, 28/12 at $K{=}1024$ and 25/11
at $K{=}2048$ ($p{=}0.02$ and $0.03$), and 25 and 18 of those wins are generations that reach the
cap under \methodseg~and stop under \methodq; at $K{=}4096$ the two are within noise on Llama-8B
and \methodseg~is ahead on Qwen-7B (16/4, $p{=}0.01$), as at the loose budgets of Table~\ref{tab:acc-math}. At $K{=}1024$ on Llama-8B every policy reaches
the cap more often than the uncapped model (100 to 131 problems against 76) and terminates
wrong less often: the accuracy eviction costs there is loops (Table~\ref{tab:gpqa}).

\begin{table}[!htbp]
\centering
\caption{\textbf{On GPQA-Diamond \methodq~is within noise of the uncapped model at every budget.} Accuracy (\%),
cap hits and terminated wrong answers on paired problems ($n{=}198$), greedy, 16,384-token cap, logical positions, compaction every $\tau{=}128$ decode steps;
H200 cards under the library stack of Table~\ref{tab:acc-math}. $^{\star}$No eviction is
budget-independent.}
\label{tab:gpqa}
\scriptsize
\setlength{\tabcolsep}{2pt}
\begin{tabular}{@{}lrrrrrrrrrrrrrrrrrr@{}}
\toprule
 & \multicolumn{9}{c}{Llama-8B} & \multicolumn{9}{c}{Qwen-7B} \\
\cmidrule(lr){2-10}\cmidrule(lr){11-19}
 & \multicolumn{3}{c}{$K{=}1024$} & \multicolumn{3}{c}{$K{=}2048$} & \multicolumn{3}{c}{$K{=}4096$} & \multicolumn{3}{c}{$K{=}1024$} & \multicolumn{3}{c}{$K{=}2048$} & \multicolumn{3}{c}{$K{=}4096$} \\
\cmidrule(lr){2-4}\cmidrule(lr){5-7}\cmidrule(lr){8-10}\cmidrule(lr){11-13}\cmidrule(lr){14-16}\cmidrule(lr){17-19}
Method & acc. & \makecell{cap\\hits} & \makecell{term.\\wrong} & acc. & \makecell{cap\\hits} & \makecell{term.\\wrong} & acc. & \makecell{cap\\hits} & \makecell{term.\\wrong} & acc. & \makecell{cap\\hits} & \makecell{term.\\wrong} & acc. & \makecell{cap\\hits} & \makecell{term.\\wrong} & acc. & \makecell{cap\\hits} & \makecell{term.\\wrong} \\
\midrule
none$^{\star}$ & 29.8 & 76 & 63 & 29.8 & 76 & 63 & 29.8 & 76 & 63 & 27.3 & 94 & 50 & 27.3 & 94 & 50 & 27.3 & 94 & 50 \\
\rowcolor{epirow}\methodq & 27.3 & 101 & 43 & 31.3 & 79 & 57 & 31.8 & 72 & 64 & 28.3 & 101 & 41 & 28.8 & 101 & 40 & 27.3 & 96 & 48 \\
\methodseg & 19.2 & 131 & 32 & 24.2 & 101 & 50 & 26.8 & 82 & 63 & 27.3 & 109 & 35 & 28.8 & 96 & 45 & 33.3 & 74 & 58 \\
RaaS & 29.8 & 100 & 39 & 32.8 & 74 & 60 & 29.3 & 72 & 68 & 25.8 & 102 & 45 & 26.8 & 85 & 61 & 23.2 & 95 & 58 \\
 
\bottomrule
\end{tabular}
\end{table}

\section{Termination}
\label{app:g-termination}

\FloatBarrier
\subsection{A generation that finds the answer and cannot stop}
\label{app:loop}

MATH-500 problem 10 asks for the least positive multiple of 30 that can be written with
only the digits 0 and 2. Both runs below are Llama-8B at $K{=}1024$ under the protocol
of Table~\ref{tab:acc-math}, on the same prompt. Under \methodseg~the model reaches
the correct answer about halfway through its generation, restates it 279 times, and
then writes the digit 2 until the 8,192-token cap, so the problem counts as a cap hit.
Under \methodq, which adds the query term and the fill, the model reaches the same answer,
checks it against the remaining candidates and stops after 3,505 tokens.

\paragraph{\methodseg, the turn into the loop.}
{\small
\begin{verbatim}
So, 2220 is the smallest four-digit number. [...] So, is 2220 the
answer? Wait, but 2220 is 2220, but 2220 is 2220, which is 2220,
but 2220 is 2220, which is 2220, which is 2220, which is 2220,
[... "which is 2220" 279 times in all ...] which is 2220. So, the
number is 2222222222222222222222222222222222222222222222222222222
[... the digit 2 until the 8,192-token cap ...]
\end{verbatim}
}

\paragraph{\methodq, the same point in the trace.}
{\small
\begin{verbatim}
Four-digit numbers: 2000, 2002, 2020, 2022, 2200, 2202, 2220, 2222.
[...] 2220 is 2220, which is 2220/30 = 74, which is integer. So, 2220
is the first four-digit number that works.

Therefore, 2220 is the smallest positive integer multiple of 30 that
can be written with only the digits 0 and 2.

**Final Answer** [...] \boxed{2220}
\end{verbatim}
}

\FloatBarrier
\subsection{Most loops are not an artifact of the length cap}
\label{app:capraise}

Under the protocol of Table~\ref{tab:acc-math}, doubling the generation cap to 16,384 tokens
at $K{=}1024$ on problems 0--99 leaves both policies' accuracy within paired noise: \methodseg~gains
two problems and \methodq~three, and neither loses any (Table~\ref{tab:capraise}). Most generations
that reach the shorter cap reach the longer one as well, 18 of 23 under \methodseg~and 10 of 16
under \methodq, so most cap hits are loops that more length would not finish; the rest end between
the two caps. The non-termination of \S\ref{sec:r-failure} therefore belongs mostly to the evicted
generation and not to the cap. An earlier grid under the rewound-position protocol of
Appendix~\ref{app:convention}, on a matched 50-problem subset, points the same way
(Table~\ref{tab:capraise-rewound}).

\begin{table}[!htbp]
\centering
\caption{\textbf{Doubling the cap leaves accuracy within noise, and most cap hits persist.}
MATH-500 problems 0--99, Llama-8B, $K{=}1024$, greedy, the protocol of
Table~\ref{tab:acc-math}. Paired: problems won / lost by the longer cap on the same problems
(exact McNemar $p \ge 0.25$).}
\label{tab:capraise}
\small
\setlength{\tabcolsep}{6pt}
\begin{tabular}{@{}lrrrrr@{}}
\toprule
 & \multicolumn{2}{c}{cap 8,192} & \multicolumn{2}{c}{cap 16,384} & \\
\cmidrule(lr){2-3}\cmidrule(lr){4-5}
Policy & acc.\ (\%) & cap hits & acc.\ (\%) & cap hits & paired \\
\midrule
\methodseg & 66.0 & 23 & 68.0 & 18 & 2 / 0 \\
\rowcolor{epirow}\methodq & 76.0 & 16 & 79.0 & 10 & 3 / 0 \\
\bottomrule
\end{tabular}
\end{table}

\begin{table}[!htbp]
\centering
\caption{\textbf{The same test under the rewound-position protocol.} Accuracy (\%) and cap hits
on a matched 50-problem MATH-500 subset at $K{=}1024$ under the two caps; rewound-position
protocol of Appendix~\ref{app:convention}, not comparable to Table~\ref{tab:acc-math}.}
\label{tab:capraise-rewound}
\small
\setlength{\tabcolsep}{6pt}
\begin{tabular}{@{}lrrrr@{}}
\toprule
 & \multicolumn{2}{c}{cap 8,192} & \multicolumn{2}{c}{cap 16,384} \\
\cmidrule(lr){2-3}\cmidrule(lr){4-5}
Policy & acc.\ (\%) & cap hits & acc.\ (\%) & cap hits \\
\midrule
\methodflat & 28.0 & 44 & 22.0 & 42 \\
\methodseg & 38.0 & 28 & 40.0 & 28 \\
\bottomrule
\end{tabular}
\end{table}

\FloatBarrier
\subsection{Protecting large value states changes neither loops nor accuracy}
\label{app:valreserve}

VaSE reports that evicting large-magnitude value states drives reasoning models into
loops that never stop \citep{vase}. If that is the mechanism behind the loops of
\S\ref{sec:r-failure}, exempting those states should lower the cap-hit rate, and a
lower cap-hit rate should buy accuracy. We add a reserve to \methodseg~that protects the
top-$r$ decode positions by value magnitude, over $r{\in}\{8,16,32,64\}$ and three
statistics of the cached value vector: its range, its variance and its $\ell_2$ norm.
All three read the cached values only, so the reserve keeps the policy on
FlashAttention. The reserve is applied as a swap rather than an addition: since
\methodseg's tier caps bind before the budget does, reserving extra slots would enlarge
the retained set and confound better tokens with more tokens, so we exchange the
lowest-scoring retained positions for the highest-magnitude evicted ones and hold the
retained count fixed.

Under the paper's protocol (logical positions, FlashAttention-2, the platform of
Table~\ref{tab:acc-math}, compaction every $\tau{=}128$ decode steps), the range reserve with $r{=}8$ matches its
unreserved baseline exactly in accuracy at $K{=}1024$ on problems 0--99 and hits the cap on two
more problems (Table~\ref{tab:valreserve-paper}).

\begin{table}[!htbp]
\centering
\caption{\textbf{A value-magnitude reserve leaves accuracy unchanged and the number of loops within noise.}
MATH-500 problems 0--99, Llama-8B, $K{=}1024$, greedy, 8,192-token cap, logical positions,
compaction every $\tau{=}128$ decode steps. Paired: problems won / lost against \methodseg~on the same problems
(exact McNemar $p{=}1.0$).}
\label{tab:valreserve-paper}
\small
\setlength{\tabcolsep}{6pt}
\begin{tabular}{@{}lrrr@{}}
\toprule
 & acc.\ (\%) & cap hits & paired \\
\midrule
\methodseg & 66.0 & 23 & -- \\
+ range reserve, $r{=}8$ & 66.0 & 25 & 6 / 6 \\
\bottomrule
\end{tabular}
\end{table}

\begin{table}[!htbp]
\centering
\caption{\textbf{A value-magnitude reserve lowers the cap-hit rate at the tight
budget and does not raise accuracy.} Accuracy and cap-hit rate (\%) on a matched
$n{=}494$ from MATH-500, exact paired McNemar against \methodseg. Absolute values are not comparable to
Table~\ref{tab:acc-math}: this comparison is internally matched but ran under the
rewound-position protocol of Appendix~\ref{app:convention}.}
\label{tab:valreserve}
\small
\begin{tabular}{@{}lrrr@{}}
\toprule
$K{=}1024$ & acc & cap-hit & paired \\
\midrule
range, $r{=}8$   & \textbf{35.6} & \textbf{61.7} & $+32/-24$, $p{=}0.35$ \\
\methodseg       & 34.0 & 64.6 & --- \\
variance, $r{=}8$ & 34.0 & 64.6 & $+29/-29$, $p{=}1.00$ \\
\midrule
$K{=}2048$ & acc & cap-hit & paired \\
\midrule
\methodseg       & \textbf{57.3} & 36.6 & --- \\
variance, $r{=}8$ & 56.9 & \textbf{36.4} & $+20/-22$, $p{=}0.88$ \\
range, $r{=}8$   & 55.9 & 38.3 & $+18/-25$, $p{=}0.36$ \\
\bottomrule
\end{tabular}

\end{table}

The full grid ran under the rewound-position protocol of Appendix~\ref{app:convention}
(Table~\ref{tab:valreserve}). There, at $K{=}1024$ the reserve lowers the cap-hit rate, as VaSE predicts, but the lower rate
does not turn into accuracy, and at $K{=}2048$ the reserve raises the cap-hit rate
slightly. No pair separates, and the direction of the
accuracy difference flips between budgets; the full grid over all twelve variants is
negative throughout. \methodseg~ships without the reserve, and the
variants are in the released code. We tested only the value-magnitude half of VaSE,
not its stochastic eviction, so this result concerns the reserve alone and not their
method.

\FloatBarrier
\subsection{\texorpdfstring{\methodname}{EpiKV} loops at the same rate on both models}
\label{app:loopshare}

Table~\ref{tab:loopshare} counts, among the problems the uncapped model solves without reaching the cap, how many
each policy turns into a loop at $K{=}1024$. \methodq's share is the same on both models;
RaaS's rises on Qwen-7B, which is where it loses most problems to the uncapped model
(63 against 42, Table~\ref{tab:termination}). The difference is not trace length: without eviction the two
models' traces are equally long (median 2,273 and 2,226 tokens on Llama-8B and Qwen-7B, with
88\% and 93\% longer than the budget).

\begin{table}[!htbp]
\centering
\caption{\textbf{\methodq's loop rate is the same on both models; RaaS's rises on Qwen-7B.}
Problems the uncapped model solves without reaching the cap on MATH-500 ($n{=}495$) that a policy turns into a
generation reaching the 8,192-token cap, at $K{=}1024$.}
\label{tab:loopshare}
\small
\setlength{\tabcolsep}{6pt}
\begin{tabular}{@{}lrr@{}}
\toprule
 & Llama-8B & Qwen-7B \\
\midrule
RaaS & 32 of 345 (9\%) & 56 of 389 (14\%) \\
\rowcolor{epirow}\methodq & 24 of 345 (7\%) & 27 of 389 (7\%) \\
\bottomrule
\end{tabular}
\end{table}

\FloatBarrier
\section{Kernel cost and serving}
\label{app:g-serving}

\FloatBarrier
\subsection{Where reading the attention weights costs}
\label{app:prefill}

Table~\ref{tab:prefill-latency} times one forward pass over a random context at batch
size one, eager attention with the weights returned against FlashAttention, keeping
only the last token's logits as a real prefill does; it is the time counterpart of
the memory curve of Figure~\ref{fig:prefillmem}. Both microbenchmarks here and the
decode-step benchmark of Table~\ref{tab:decode-step} ran on one exclusive
A100-40GB under transformers 4.57.1, each attention mode in its own process; the
memory curve of Figure~\ref{fig:prefillmem} ran on an A100-80GB under transformers
5.2.0. The same measurements on an L40S show the same pattern and are released
with the code.

\begin{table}[!htbp]
\centering
\caption{\textbf{Reading the attention weights makes a prefill pass 1.3$\times$ slower
at 1,024 tokens and 1.7$\times$ at 2,048, and the pass fails at 4,096 on a 40\,GB
card; the FlashAttention pass fits 32,768.} One forward pass of Llama-8B over a random
context of the given length, batch size one, bf16, one A100-40GB, transformers
4.57.1; time is the median of three passes after one warm-up pass and peak memory is
the allocator's peak.}
\label{tab:prefill-latency}
\small
\setlength{\tabcolsep}{6pt}
\begin{tabular}{@{}rrrrr@{}}
\toprule
 & \multicolumn{2}{c}{attention weights read (eager)} & \multicolumn{2}{c}{FlashAttention} \\
\cmidrule(lr){2-3}\cmidrule(lr){4-5}
context (tokens) & peak GB & time (s) & peak GB & time (s) \\
\midrule
512 & 15.6 & 0.06 & 15.1 & 0.06 \\
1,024 & 17.3 & 0.12 & 15.3 & 0.09 \\
2,048 & 24.1 & 0.29 & 15.7 & 0.17 \\
4,096 & OOM & -- & 16.4 & 0.32 \\
8,192 & OOM & -- & 17.8 & 0.67 \\
16,384 & OOM & -- & 20.5 & 1.47 \\
32,768 & OOM & -- & 26.1 & 3.57 \\
65,536 & OOM & -- & OOM & -- \\
 
\bottomrule
\end{tabular}

\end{table}

\begin{table}[!htbp]
\centering
\caption{\textbf{Returning the attention weights costs nothing when one request
decodes at a time up to 8,192 cached tokens, and up to 2.2 times the step time under batching, where it also runs
out of memory first.} Time of one decode step of Llama-8B against a cache of the given
length, eager attention with the weights returned against FlashAttention; one A100-40GB,
transformers 4.57.1, each attention mode in its own process, median of five steps
after one warm-up.}
\label{tab:decode-step}
\small
\setlength{\tabcolsep}{6pt}
\begin{tabular}{@{}rrrrr@{}}
\toprule
batch & cache (tokens) & eager (ms) & FlashAttention (ms) & ratio \\
\midrule
1 & 1,024 & 35.9 & 37.9 & 0.95 \\
 & 2,048 & 35.9 & 37.8 & 0.95 \\
 & 4,096 & 35.9 & 37.9 & 0.95 \\
 & 8,192 & 36.0 & 37.9 & 0.95 \\
 & 16,384 & 50.4 & 40.2 & 1.25 \\
\addlinespace
8 & 1,024 & 36.9 & 36.0 & 1.03 \\
 & 2,048 & 50.9 & 35.9 & 1.42 \\
 & 4,096 & 80.1 & 36.2 & 2.21 \\
 & 8,192 & OOM & 52.8 & -- \\
\addlinespace
32 & 1,024 & 83.8 & 39.0 & 2.15 \\
 & 2,048 & OOM & 56.0 & -- \\
 
\bottomrule
\end{tabular}

\end{table}

\paragraph{Every policy at equal load.} Table~\ref{tab:time-aime} runs every policy
end to end with one process per card on the same AIME-2024 problems. The decode rate
does not split by attention kernel. The eager methods are among the fastest rows on
both models, and on Llama-8B the slowest rows are FA2 policies that rescore the whole
cache at every step. Most capped caches decode faster than no eviction at this
generation length, because the uncapped cache keeps growing; on Llama-8B the exceptions
are the two rescoring policies and RaaS. Peak memory in this regime is set by the weights and
the cache, not by the method: the rows sit within about a gigabyte of one another,
and no eviction is the lowest of them, since a compaction briefly holds two copies
of the cache and the eager kernels hold their attention maps. \methodq's query term costs nothing measurable in decode rate (26.9 against 26.2
tokens/s for \methodseg) and 0.6\,GB of memory, the relevance matrix it holds once per compaction.

\begin{table}[!htbp]
\centering
\caption{\textbf{At batch size one the decode rate is set by what a policy does between
forward passes, not by the attention kernel.} AIME-2024 (30 problems), $K{=}8192$,
16,384-token cap, greedy, one process per A100-80GB (transformers 4.57.1, torch 2.7,
FlashAttention-2 2.8.3). tok/s: generated tokens over wall time, summed over the problems;
cap: problems reaching the cap; GB: mean peak GPU memory. Accuracy on 30 problems ($\pm$16
points) only ties each row to its run. The \methodq~row ran on Llama-8B only, on one
exclusive A100.}
\label{tab:time-aime}
\small
\setlength{\tabcolsep}{3.2pt}
\begin{tabular}{@{}llrrrrrrrrrr@{}}
\toprule
 & & \multicolumn{5}{c}{Llama-8B} & \multicolumn{5}{c}{Qwen-7B} \\
\cmidrule(lr){3-7}\cmidrule(lr){8-12}
Method & Kernel & acc & cap & tok/s & s/prob & GB & acc & cap & tok/s & s/prob & GB \\
\midrule
No eviction & FA2 & 33.3 & 16 & 19.9 & 606 & 16.5 & 33.3 & 19 & 20.9 & 572 & 14.9 \\
\rowcolor{epirow}\methodq~($\tau{=}128$) & FA2 & 33.3 & 15 & 26.9 & 437 & 17.6 & -- & -- & -- & -- & -- \\
\methodseg~($\tau{=}128$) & FA2 & 26.7 & 18 & 26.2 & 473 & 17.0 & 36.7 & 17 & 25.9 & 453 & 15.0 \\
\methodseg~($\tau{=}1$) & FA2 & 30.0 & 13 & 27.2 & 427 & 17.0 & 33.3 & 18 & 20.8 & 566 & 15.0 \\
raw score & FA2 & 33.3 & 16 & 16.1 & 738 & 17.4 & 36.7 & 18 & 27.4 & 430 & 15.2 \\
KV-key & FA2 & 33.3 & 16 & 17.3 & 679 & 17.4 & 40.0 & 16 & 27.8 & 419 & 15.2 \\
RaaS & eager & 26.7 & 19 & 18.1 & 688 & 17.6 & 40.0 & 17 & 19.9 & 612 & 15.4 \\
H2O & eager & 30.0 & 18 & 27.2 & 456 & 17.6 & 43.3 & 16 & 29.2 & 408 & 15.4 \\
R-KV & eager & 26.7 & 19 & 29.4 & 428 & 17.6 & 40.0 & 15 & 22.4 & 529 & 15.4 \\
LongFlow & eager & 30.0 & 17 & 25.8 & 486 & 17.6 & 36.7 & 16 & 29.5 & 413 & 15.4 \\
ThinKV & eager & 26.7 & 18 & 29.7 & 427 & 16.9 & 36.7 & 18 & 30.4 & 408 & 15.1 \\
\bottomrule
\end{tabular}

\end{table}

\FloatBarrier
\subsection{\texorpdfstring{\methodname}{EpiKV} inside vLLM}
\label{app:vllm}

\paragraph{The eviction port.} Both policies run inside vLLM~0.11.2 on stock kernels
(Table~\ref{tab:vllm-evict}). Every $\tau{=}128$ decode steps, when a request's logical
length exceeds $K$, the port scores the request's cached positions, compacts its paged
cache in place, and tells the scheduler's block allocator which blocks the request
still owns; the attention kernels and the engine loop are unchanged. The port sends the
harness's exact prompt token ids, system instruction included. At the first compaction of MATH-500 problem 1 we saved, from both
engines, the Band-A hidden states, the recomputed queries, the cached keys and the
query-term relevance at layer 11. The cached keys match to within 1\% (relative $\ell_1$
difference 0.8\%, cosine 0.99997), the hidden states and query norms to within 3 and
6\%, and the relevance vectors have cosine 0.993. Their rankings agree only in part,
however: the rank correlation is 0.38, and 65\% of the 512 most relevant positions are
shared. The two engines therefore keep partly different tokens, and they reach the same
accuracy within paired noise.

\paragraph{The cost of the score.} We register forward hooks on layers 10 and 21 of
vLLM~0.11.2's decode path (A100-80GB, 500-token prompts, 512 decode tokens, three runs
per configuration) and compute the rolling detrended score online. Under continuous
batching the per-layer hidden-state tensor concatenates all sequences in the step, so
per-sequence changes are recovered exactly from the scheduler's
\texttt{query\_start\_loc} offsets. The hooked values match HuggingFace's
\texttt{output\_hidden\_states}, and computing the score online matches computing it
offline from the same saved values (Table~\ref{tab:vllm}).

\begin{table}[!htbp]
\centering
\caption{\textbf{Computing the score online inside vLLM costs a modest share of
decode throughput at every concurrency.} Isolated by a paired comparison, the
engine with and without the hooks back to back, three runs each. Prefill cost is at
most 1.7\%; the tracker state is a constant of about 9\,MB. The hooked deltas
match HuggingFace's within 1.1\% relative error, the bf16 accumulation floor, and
offline recomputation reproduces the live scores to $10^{-6}$. vLLM's own
eager-versus-CUDA-graph gap is 7--39\%, shrinking with concurrency.}
\label{tab:vllm}
\normalsize
\begin{tabular}{@{}lrrrr@{}}
\toprule
Concurrency & 1 & 8 & 32 & 64 \\
\midrule
Decode-throughput cost & 13.9\% & 9.2\% & 7.3\% & 11.1\% \\
\bottomrule
\end{tabular}

\end{table}

\begin{figure}[h]
\centering
\includegraphics[width=\textwidth]{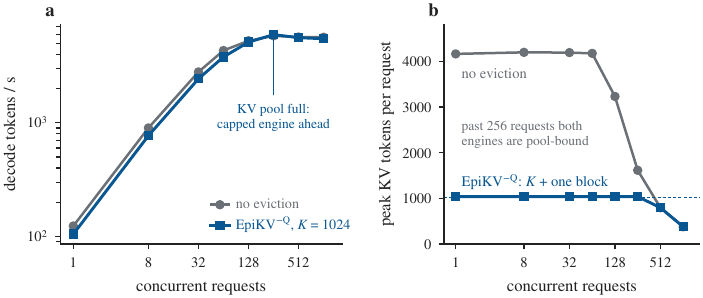}
\caption{\textbf{Capping the cache costs a modest share of decode throughput while memory is
plentiful and nothing once it is not.} The same vLLM engine with and without \methodseg~eviction
at $K{=}1024$, one H100, every request decoding 4,096 tokens. \textbf{(a)} Decode tokens per
second; from 128 concurrent requests the uncapped engine runs out of KV memory and queues.
\textbf{(b)} Peak KV-cache tokens held per request: $K$ plus one block for the capped engine;
the uncapped value falls once requests wait for blocks.}
\label{fig:vllm-tp}
\end{figure}

Three points of deployment context. First, hooks require eager execution, and a
CUDA-graph-compatible integration would fold the score into the graph
(Table~\ref{tab:vllm} gives the eager-versus-graph gap). Second, the
claim in \S\ref{sec:r-eng} that no vLLM decode path exposes attention weights
rests on a scan of the installed source: the sole \texttt{output\_attentions}
occurrence is in a legacy vision encoder,
\texttt{model\_executor/models/swin.py}. Third, at the 4,096-token decode
length of Figure~\ref{fig:vllm-tp} the capped engine's own pool binds only past 256 requests; the
throughput case for capping is at the 16,384- to 65,536-token lengths of
Figure~\ref{fig:prefillmem}, which we did not run inside vLLM.

\FloatBarrier
\subsection{How many requests fit, with and without a budget}
\label{app:throughput}

\begin{figure}[h]
\centering
\includegraphics[width=0.85\linewidth]{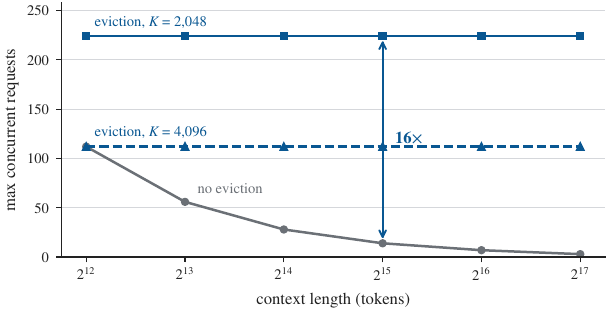}
\caption{\textbf{A fixed cache budget holds the number of concurrent requests flat
as traces grow; without eviction it falls with the context length.} Maximum
concurrent requests on an 80\,GB GPU against context length, from the KV arithmetic
above; the marked gap is the context-to-budget ratio.}
\label{fig:batch}
\end{figure}

Figure~\ref{fig:batch} projects, from the per-token KV-cache size, the maximum number
of concurrent requests that fit on an 80\,GB GPU with and without a budget; the
measured counterpart is Figure~\ref{fig:vllm-tp}. For Llama-8B
(32 layers, 8 KV heads, head dimension 128, 16-bit entries) the KV cache costs
$2 \times 32 \times 8 \times 128 \times 2$ bytes $= 0.125$\,MiB per token. At a
32{,}768-token context, a full (no-eviction) cache is therefore 4\,GiB per
request, whereas a cache capped at a 2{,}048-token budget is 0.25\,GiB. With
${\approx}56$\,GiB of the 80\,GB available for KV after model weights and runtime
workspace, this yields $\lfloor 56/4 \rfloor = 14$ concurrent requests without
eviction versus $\lfloor 56/0.25 \rfloor = 224$ with the capped cache, a
$16\times$ ratio equal to the context-to-budget ratio.

\end{document}